\pdfoutput=1
\documentclass[11pt]{article}

\usepackage[final]{acl}
\usepackage{times}
\usepackage{latexsym}

\usepackage[T1]{fontenc}
\usepackage[utf8]{inputenc}

\usepackage{microtype}

\usepackage{graphicx}
\usepackage{graphics}
\usepackage{caption}
\usepackage{subfigure}

\usepackage{booktabs}       % professional-quality tables
\usepackage{amsfonts}       % blackboard math symbols
\usepackage{nicefrac}       % compact symbols for 1/2, etc.
\usepackage{microtype}      % microtypography
\usepackage{amsmath}
\usepackage{multirow}
\usepackage{soul}

\usepackage{svg}
\usepackage{float}
\usepackage{color}
\usepackage{xcolor}
\usepackage{array}

\title{A Generative Language Model for Few-shot Aspect-Based Sentiment Analysis}

\author{Ehsan Hosseini-Asl, Wenhao Liu, Caiming Xiong \\
        Salesforce AI Research \\ \texttt{\{ehosseiniasl,wenhao.liu,cxiong\}@salesforce.com} }
\begin{document}
\maketitle

\begin{abstract}
Sentiment analysis is an important task in natural language processing.
In recent works, pre-trained language models are often used to achieve state-of-the-art results, especially when training data is scarce. It is common to fine-tune on the downstream task, usually by adding task-specific layers on top of the model.
In this paper, we focus on aspect-based sentiment analysis, which involves extracting aspect term, category, and predicting their corresponding polarities. In particular, we are interested in few-shot settings.
We propose to reformulate the extraction and prediction tasks into the sequence generation task, using a generative language model with unidirectional attention (GPT2 is used unless stated otherwise). This way, the model learns to accomplish the tasks via language generation without the need of training task-specific layers. 
Our evaluation results on the single-task polarity prediction show that our approach outperforms the previous state-of-the-art (based on BERT) on average performance by a large margins in few-shot and full-shot settings. More importantly, our generative approach significantly reduces the model variance caused by  low-resource data.
We further demonstrate that the proposed generative language model can handle joint and multi-task settings, unlike previous work.
We observe that the proposed sequence generation method achieves further improved performances on polarity prediction when the model is trained via joint and multi-task settings. Further evaluation on similar sentiment analysis datasets, SST-2, SST-5 and OOS intent detection validates the superiority and noise robustness of generative language model in few-shot settings.

\end{abstract}

\section{Introduction}
\label{sec:intro}
Sentiment analysis~\cite{pang2002thumbs,turney2002thumbs, chevalier2006effect,bastan-etal-2020-authors} aims at detecting the overall polarity of a user generated text, which describes the user opinion for an entity. However, user may express opinions about an entity at different granularity. For example, a user may give an overall rate about a restaurant service, and then explains fine-grained review about specific aspects, such as food quality, waiting time, waitress service, environment, etc. Aspect-based sentiment analysis task~\cite{pontiki-etal-2014-semeval, pontiki2016semeval} aims at addressing this problem, where user sentiment is annotated at coarse and fine-grained levels. Moreover, user can express conflicting opinions for different aspects of an entity. %For example 

Traditionally, neural-based models are employed as a single-task model for aspect-based sentiment analysis (ABSA) task, 
% employed as a single task model, 
% i.e. prediction polarity given the aspect term or aspect term extraction only.
similar to Machine Reading Comprehension task (MRC)~\cite{rajpurkar2016squad}.
% \hl{[TODO: More edits]}. 
For example, a pre-trained BERT language model is fine-tuned for ABSA term polarity prediction (single-task) as a classifier. In this approach, a task-specific layer is fine-tuned for each downstream task, such as a layer for aspect term polarity classification, and a different layer for aspect term span extraction~\cite{xu2019bert}. 

Recently, generative language models with unidirectional self-attention, which are pre-trained by causal language modeling loss (predicting next word given the history), have shown promising performance when fine-tuned on the downstream tasks (GPT2)~\cite{radford2018improving}. Using this approach, the language model learns the downstream task as language generation, where the task is represented as a serialized text. Moreover,~\citet{brown2020language} proposed GPT3, a large-scale generative language model with few-shot ability. GPT3 learns to solve the downstream task by conditioning on few examples in the prompt, without any parameter update (in-context learning). 

Motivated by the ability of the pre-trained generative language model (GPT2) for solving the downstream tasks in a generative manner, 
% and few-shot performance (GPT3), 
we propose a generative language model for ABSA task. 
The evaluation results indicate that the proposed approach achieves better performance with significantly lower variance compared to the previous state-of-the-art models 
(which are based on BERT pre-trained model) on few-shot and full-shot settings, for single-task polarity prediction of aspect term and aspect category. 
For example, using $1\%$ ($20$ examples) of training data on restaurant domain for aspect term polarity prediction task, our proposed GPT2 model outperforms BERT-PT~\cite{xu2019bert} by $9$ points on average accuracy and reduced standard deviation by $6.2$ points, as shown in Figure~\ref{fig:single_task_polarity}(a). Moreover, when fine-tuned on multiple tasks, such as aspect term extraction, term polarity, aspect category detection, and category polarity, the proposed model improved single-task performance, such as aspect term extraction (measured by F1 score).
\footnote{Code is available at \url{https://github.com/salesforce/fewshot_absa}}

The contributions of our proposed generative language model are, 
% \wl{merge bullet 1, 3, 4 into one point, about new task formulation and the improvement}
% \eh{fixed}

\begin{itemize}
    \item A robust generative model on few-shot aspect-based sentiment analysis by reformulating the task as language generation. This allows us to use uni-directional language model with no additional head for the downstream tasks, which outperforms the previous state-of-the-arts on average performance by a large margin, with no additional pretraining on out-of-domain data (such as BERT-PT~\cite{xu2019bert}).
    % and reducing the variance due to random noise. 
    % \item Reformulating aspect-based sentiment analysis as language generation. This allows us to use uni-directional language model with no additional head for the downstream tasks.
    \item Our proposed generative model reduces variance in polarity prediction, caused by low resource data and random noise, in all few and full-shot settings by large value. 
    % with no additional pretraining on out-of-domain data (such as BERT-PT~\cite{xu2019bert}).
    % \item Improving full-shot performance on all polarity prediction single-tasks, without additional pre-training on out-of-domain data (such as BERT-PT~\cite{xu2019bert}).
    % \item A robust generative model on few-shot setting, which outperforms the previous state-of-the-arts by a large margin. 
    % \item We present a new method for creating the input sequence during training to improve few-shot performance of the generative language model (GPT2). 
    \item 
    % Evaluation indicates that using 
    % It is shown that 
    Joint and multi-task training can further improve the single-task few-shot performances, such as aspect term extraction. 
    \item More evaluation on similar 
    % \wl{typo - similar} 
    sentiment analysis tasks (SST-2, SST-5, OOS intent detection) provides further evidence of the superiority and robustness of generative language model.
\end{itemize}

In the next sections, we discuss the proposed model and presents the evaluation results. In section~\ref{sec:related}, the previous state-of-the-arts are described. Section~\ref{sec:model} explains the task of aspect-based sentiment analysis (ABSA) (section~\ref{subsec:absa_task}) followed by reformulating ABSA task as language generation~(section~\ref{subsec:generative_lm}). In section~\ref{sec:exp}, the evaluation results for single, joint and multi-task settings are presented for SemEval14~\cite{pontiki-etal-2014-semeval} and SemEval16~\cite{pontiki2016semeval} 
% datasets. 
and SST-2, SST-5 and OOS intent detection datasets.

\section{Related Works}
\label{sec:related}
Sentiment analysis is characterized by three categorizes, i.e. document, sentence, and aspect level~\cite{liu2012sentiment, liu2012survey, cambria2012sentic}. In this section, we review the previous models developed for aspect-based sentiment analysis (ABSA)~\cite{hu2004mining}. 

Earlier works on ABSA task focused on developing feature engineered models~\cite{samha2014aspect}.~\citet{xu2018double} proposed a model based on using convolutional neural network (CNN) for aspect term extraction task only. The approach uses two types of pre-trained embeddings, a general-purpose embedding and a domain-specific one. Then, a softmax classification layer is used to classify each word to identify aspect term start and end positions, or non-related words.

% \paragraph{RNN-based approaches:}
\citet{li2019exploiting} proposed Multi-granularity Alignment Network (MGAN), a coarse-to-fine approach for single-task aspect term polarity prediction using recurrent neural network (RNN)~\cite{hochreiter1997long}. They defined aspect category as coarse-level and aspect term as fine-level sentiments, and further leveraged high-resource out-of-domain data for pre-training. This way, the knowledge is transferred from coarse-grain domains (single-opinion prediction) to multi-grain domains (ABSA task). 
%, they pretrained a model for fine-grained sentiment analysis (ABSA) as downstream task.

% \paragraph{BERT-based approaches} 
With the advent of BERT~\cite{devlin2018bert} as a pre-trained bidirectional language model, which presents a powerful contextualized word representation for the language understanding downstream tasks, several models are proposed for ABSA task using BERT as feature extraction.~\citet{xu2019bert} defined ABSA task as question answering~\cite{rajpurkar2016squad}, named Review Reading Comprehension (RRC), and used BERT as the base model, with separate heads for aspect term extraction (as span extraction) and term polarity prediction. To enhance RRC performance, they introduced a post-training algorithm, which additionally pre-train the model on out-of-domain data from Amazon and Yelp review datasets, and additionally on MRC question answering dataset~\cite{rajpurkar2016squad}. These result in additional training set of $1,151,863$ for laptop domain, $2,677,025$ more examples for restaurant domain, and $87,599$ training examples from MRC dataset. 

\citet{karimi2020improving} proposed an approach based on conditional random field (CRF)~\cite{lafferty2001conditional}, combined with BERT for aspect term extraction and term polarity prediction tasks. Two modules are employed for improving aspect term extraction and term polarity prediction of BERT model. First, a parallel approach is used which combines predictions for aspect term and polarity from last four layers of BERT in parallel. Moreover, a hierarchical aggregation module is also examined, where predictions of previous layers of BERT are fed into the next layer.~\citet{reddy2020does} combines GLOVE pre-trained embedding~\cite{pennington2014glove} with deep contextualized representation of BERT to enhance the representation of word vectors for predicting aspect term polarity. 
% semantic (aspect term polarity prediction task). 
The proposed BERT-IL model predicts aspect term polarity by learning a similarity between GLOVE vector of aspect term and its contextualized representation extracted from BERT. First, the aspect term representations are extracted from multiple layers of BERT, and fed into a self-attention layer. Finally, it is further fine-tuned on ABSA task for performance improvement.
% (BERT-IL Finetuned). 
\citet{liu-etal-2021-solving} proposed a model based on BART~\cite{lewis-etal-2020-bart} for aspect category detection. They rank all aspect categories with different polarities and select the pair with highest score. 
In section~\ref{sec:exp}, evaluation of our proposed generative language model are compared with the recent BERT-based models. 

% The evaluation of 

% module are trained on top of BERT for 

% proposed a post-trained BERT model (BERT-PT) to solve the task. 
% By additionally pretraining BERT on out-of domain data from Yelp and Amazon dataset, they further finetuned BERT on auxiliary task of review reading comprehension, similar to MRC~\cite{rajpurkar2016squad}. 

\vspace{-2mm}
\section{Model}
\label{sec:model}
\vspace{-2mm}
This section describes aspect-based sentiment analysis task (ABSA), the proposed generative language model approach, details of the datasets, model training, and evaluation metrics.
% training details, dataset details, and evaluation metrics. 
% followed by modeling it as a causal language modeling. 

\subsection{Aspect Based Sentiment Analysis}
\label{subsec:absa_task}
% \vspace{-2mm}
Aspect-based sentiment analysis (ABSA) is similar to sentiment analysis, in the sense that the task is to predict the polarity of an entity in a sentence. However, it is different, since the goal is to predict fine-grained sentiment of multiple aspect terms and categories of an entity. The task was first introduced in Semantic Evaluation Challenge (SemEval14)~\cite{pontiki-etal-2014-semeval}. It was then extended in SemEval16 challenge~\cite{pontiki2016semeval}. 
% The two datasets, i.e. SemEval14 and SemEval16,
The challenges comprise of two domains, restaurant and laptop, where each domain spans over four sub-tasks (SB1-4).

% \begin{table}[htb!]
%     \centering
%     \begin{tabular}{|c|c|c|c|c|}
%         \hline
%         Dataset & Domain & Train & Test & Total \\
%         \hline
%         \multirow{2}{*}{SemEval14} & Restaurant & 3041 & 800 & 3841 \\
%          & Laptop & 3045 & 800 & 3845 \\
%          \hline
%          \multirow{2}{*}{SemEval16} & Restaurant & 2000 & 676 & 2676 \\
%           & Laptop & 2500 & 808 & 3308 \\
%           \hline
%     \end{tabular}
%     \caption{dataset distribution}
%     \label{tab:dataset}
% \end{table}

\paragraph{Aspect Term Extraction (SB1)}
For a given review sentence, this sub-task is about predicting all aspects terms (word span) that opinions are expressed. It requires that all aspect terms to be predicted, including those which no opinion is expressed (neutral sentiment). This sub-task (AE) corresponds to sub-task 1 (SB1) - single sentence -- slot 2 in SemEval16 challenge, named as opinion target expression (OTE)~\cite{pontiki2016semeval}.

\paragraph{Aspect Term Polarity (SB2)}
For a given review sentence and an aspect term, the goal is to predict the polarity of the expressed opinion (\verb|positive|, \verb|negative|, \verb|neutral|, \verb|conflict|). This sub-task corresponds to SB1-Slot3 in SemEval16 challenge.

\paragraph{Aspect Category Detection (SB3)}
Given a set of pre-defined aspect categories 
% \wl{this is only for the restaurant domain, right?} \eh{No, both domains, see Fig1(d)} 
(e.g. \verb|PRICE|, \verb|FOOD|, \verb|SERVICE|, \verb|AMBIENCE|, \verb|ANECDOTE/MISCELLANEOUS|), the goal is to predict all categories that an opinion is expressed about. This sub-task corresponds to SB1-Slot1 (single-sentence) in SemEval16 challenge, where the category is defined as the pair of entity and attribute, e.g. \verb|RESTAURANT#PRICE|, \verb|FOOD#QUALITY|, \verb|LAPTOP#GENERAL|, \verb|LAPTOP#PRICE|. Please refer to Table~\ref{tab:appendix_aspect_category} in the appendix for the full list of categories for laptop and restaurant domains.
% \wl{this is still just listing categories from the restaurant domain} \eh{Table 6 added in appendix}

\paragraph{Aspect Category Polarity (SB4)}
Given a review sentence and a category, the goal is to predict the sentiment of the category (\verb|positive|, \verb|negative|, \verb|neutral|, \verb|conflict|). This sub-task corresponds to SB1-Slot3 in SemEval16~\cite{pontiki2016semeval}.

\subsection{Generative Language Modeling}
\label{subsec:generative_lm}
ABSA task comprises of four sub-tasks: aspect term extraction, aspect category detection, and aspect term and category polarity predictions. The dominant approach for solving ABSA task is to train separate classifiers for each sub-task~\cite{xu2019bert}. In this paper, we propose to solve all sub-tasks using a single auto-regressive (generative) language model, either using single-task or joint-task training.

\subsubsection{Language model}
\label{subsec:lm}
The goal of generative language modeling is to learn data distribution $p(x)$, where $x=(x_{1},\ldots,x_{n})$ is a sequence of $n$ symbols. In order to model $p(x)$, the language model factorizes the distribution of a single sequence $p(x)$ using the chain rule of probability~\cite{bengio2003neural}, and training a neural network, which is parameterized by $\theta$, by minimizing the negative log-likelihood,

\begin{align}
    p_{\theta}(x) & = \prod_{t=0}^{n} p_{\theta}(x_{t}|x_{<t})
    \label{eq:lm_prob} \\
    \mathcal{L}_{D} & = -\sum_{k=1}^{K} \sum_{t=1}^{n} \log p_{\theta}(x^{k}_{t}|x^{k}_{<t})
    \label{eq:lm_loss}
\end{align}

During inference, the generative model sequentially generates tokens by conditioning on the input example $x^{k}$, and the past generated tokens.

\subsection{ABSA task as generative language modeling}
\label{subsec:absa=_lm}

Each ABSA task training example, $x^{k}$, contains a sentence $S^{k}$, $I$ pairs  of aspect term and term polarity, and $J$ pairs of aspect category and category polarity, 
\begin{align}
    T^{k} & =\{TP^{k}_{i}=(t^k_{i}, pt^{k}_{i}) ; i\in I\} \label{eq:term_pair} \\
    % \sum_{i=1}^{I}(T^{k}_{i}, {Pt}^{k}_{i}) \\
    % S^{k} & = \sum_{j=1}^{J}(C^{k}_{j}, {Pc}^{k}_{j})
    C^{k} & =\{CP^{k}_{j}=(c^k_{j}, pc^{k}_{j}) ; j\in J\} \label{eq:category_pair} 
\end{align}
where $t^{k}_{i}$, $pt^{k}_{i}$, and $TP^{k}_{i}$ are $i$-th aspect term, term polarity, and their pair. Moreover, $c^{k}_{j}$ and $pc^{k}_{j}$, and $CP^{k}_{j}$ are $j$-th aspect category, category polarity, and their pair of $k$-th sentence. 

% $T^{k}$ $T_{k}=\sum_{i=1}^{I}(T_{k}^{i})$, term polarities $P^{T}_{k}=\sum_{i=1}^{I}(P_{k}^{i})$, aspect categories $\sum_{j=1}^{J}C_{k}$, and category polarity $\sum_{j=1}^{J}P^{C}_{k}$. 

\subsubsection{Single-Task Polarity Prediction} 
% \wl{in this section, we should use specific examples from the dataset wherever possible. the abstract equation form doesn't help the reader very much, and it's easy to be lost in wondering ``what does the model take as input and what is it trained to generate"?} 
% \eh{see highlighted added description with reference to appendix with tables of input sequence for training and inference}
\label{subsec:single_task}
This task consists of predicting the polarity of aspect terms or aspect categories only (named as SB2 and SB4 in section~\ref{subsec:absa_task}). To generate polarity during the inference, the input to the generative language model (LM) comprises of $k$-th sentence and the corresponding aspect term or category,
\begin{align}
     pt^{k}_{i} & = LM_{term}(S^{k}, t^{k}_{i}) \label{eq:single-task_term}\\
     pc^{k}_{j} & = LM_{category}(S^{k}, c^{k}_{j}) \label{eq:single-task_category}
\end{align}
where $LM_{term}$ refers to a model that trained on aspect term dataset, and $LM_{category}$ refers to aspect category dataset, respectively. The details of training language model are described in section~\ref{subsec:trainig}. Moreover, the details of input sequence formulation during training and inference are presented in Appendix~\ref{appendix_input_rep} and Tables~\ref{tab:method} and~\ref{tab:input_sequence_examples}.

\subsubsection{Joint and Multi-Task Prediction} 
\label{subsec:joint_task}
This task includes generating pairs of aspect term and term polarity, or pairs of aspect category and their polarity.
% pairs, respectively.
To jointly generate aspect terms and their polarities, the model input relies on the review sentence $S^{k}$ only, and the model outputs all aspect term and polarity pairs in token-by-token (auto-regressive) generation,

\begin{align}
     T^{k} & = LM_{term}(S^{k}) \label{eq:joint-task_term}\\
     C^{k} & = LM_{category} (S^{k}) \label{eq:joint-task_category}
\end{align}
where $T^{k}$ is the set of aspect term and polarity pairs, Eq.~(\ref{eq:term_pair}), and $C^{k}$ is the set of aspect category and polarity pairs, Eq.~(\ref{eq:category_pair}). 
The same method in joint-task prediction can be used to generate all pairs of aspect term and aspect category, i.e. multi-task prediction,
\begin{equation}
    [T^{k}; C^{k}] = LM_{multi}(S^{K})
    \label{eq:multi-task}
\end{equation}

In this case, during training, the model learns to generate $I$ pairs of aspect term and $J$ pairs of aspect category via language model training,~Eq.~(\ref{eq:lm_prob}). 
% During inference, given the input sentence 

% \begin{equation}
%      \sum_{i=1}^{I}(T_{i}, P^{T}_{i}) \sum_{j=1}^{J}(C_{j}, P^{C}_{j})= CLM(S)
% \end{equation}

\subsubsection{Training}
\label{subsec:trainig}
A training sequence for solving each sub-tasks (SB1-4) of section~\ref{subsec:absa_task}, consists of the review sentence, concatenated by the corresponding aspect term/category and its polarity. 
% For example, for training language model for SB2, 
For example, in training $LM_{term}$ for predicting aspect term polarity (Eq.~\ref{eq:single-task_term}) and joint-task prediction of aspect term and polarity (Eq.~\ref{eq:joint-task_term}), the training sequence comprises of the review sentence concatenated by aspect terms and their polarities, $x^{k}=[S^{k}; T^{k}]$. 
% This training example is used for single-task term polarity (Eq.~\ref{eq:single-task_term}) and joint-task training of aspect term and polarity tasks (Eq.~\ref{eq:joint-task_term}), namely $LM_{term}$. 
Respectively, $x^{k}=[S^{k}; C^{k}]$ is used for training $LM_{category}$, as mentioned in Eq.~(\ref{eq:single-task_category}) and (\ref{eq:joint-task_category}). For more details on input sequence representation, see Appendix~\ref{appendix_input_rep}, Tables~\ref{tab:method} and~\ref{tab:input_sequence_examples}. 

In order to train $LM_{term}$, the model can be trained on different training sequences, where the review sentence $S^{k}$ needs to only be concatenated with a single pair of aspect term and polarity. In this case, multiple training sequences are created for the $k$-th sentence, i.e. $\{x^{k}_{i}=[S^{k}; TP^{k}_{i}]; i\in I\}$. 
We will present 
An ablation study on these two methods of sequence creating for the language model training, and its effect on few-shot and full-shot performances, are presented in Appendix~\ref{sec:appendix_ablation_input_formatting_methods} and Figure~\ref{fig:appendix_ablation_input_format}.
% section~\ref{exp:ablation}. 

\begin{table}[htb!]
    \centering
    % \small
    \scriptsize
    \begin{tabular}{|c|c|c|c|c|}
        \hline
        Dataset & Domain & Train & Dev & Test %& Total \\
        \\
        \hline
        \multirow{2}{*}{SemEval 14} & Restaurant & 3041 & -  & 800 
        %& 3841 
        \\
         & Laptop & 3045 & - & 800 %& 3845 
         \\
         \hline
         \multirow{2}{*}{SemEval 16} & Restaurant & 2000 & - & 676 
         %& 2676 
         \\
          & Laptop & 2500 & - & 808 \\
          %& 3308 \\
        \hline
         SST-2 & Movie &  66749 & 872 & 1821 \\
         %& 69442 \\
         \hline
         SST-5 & Movie & 8544 & 1101 & 2210 \\
         %& 11855 \\
         \hline
         OOS & Misc. & 15100 & 3100 & 4500 \\
         %& 22700 \\
          \hline
    \end{tabular}
    \vspace{-2mm}
    \caption{Dataset distribution 
    % \wl{table 1 is a bit too wide} \eh{fixed}
    }
    \label{tab:dataset}
    \vspace{-3mm}
\end{table}

\begin{figure*}[htb!]
    
    % \begin{subfigure}[0.4\linewidth]
    \centering
    \subfigure[
    % semeval 14, restaurants term polarity (SB1)
    ]{\includegraphics[width=0.3\linewidth]{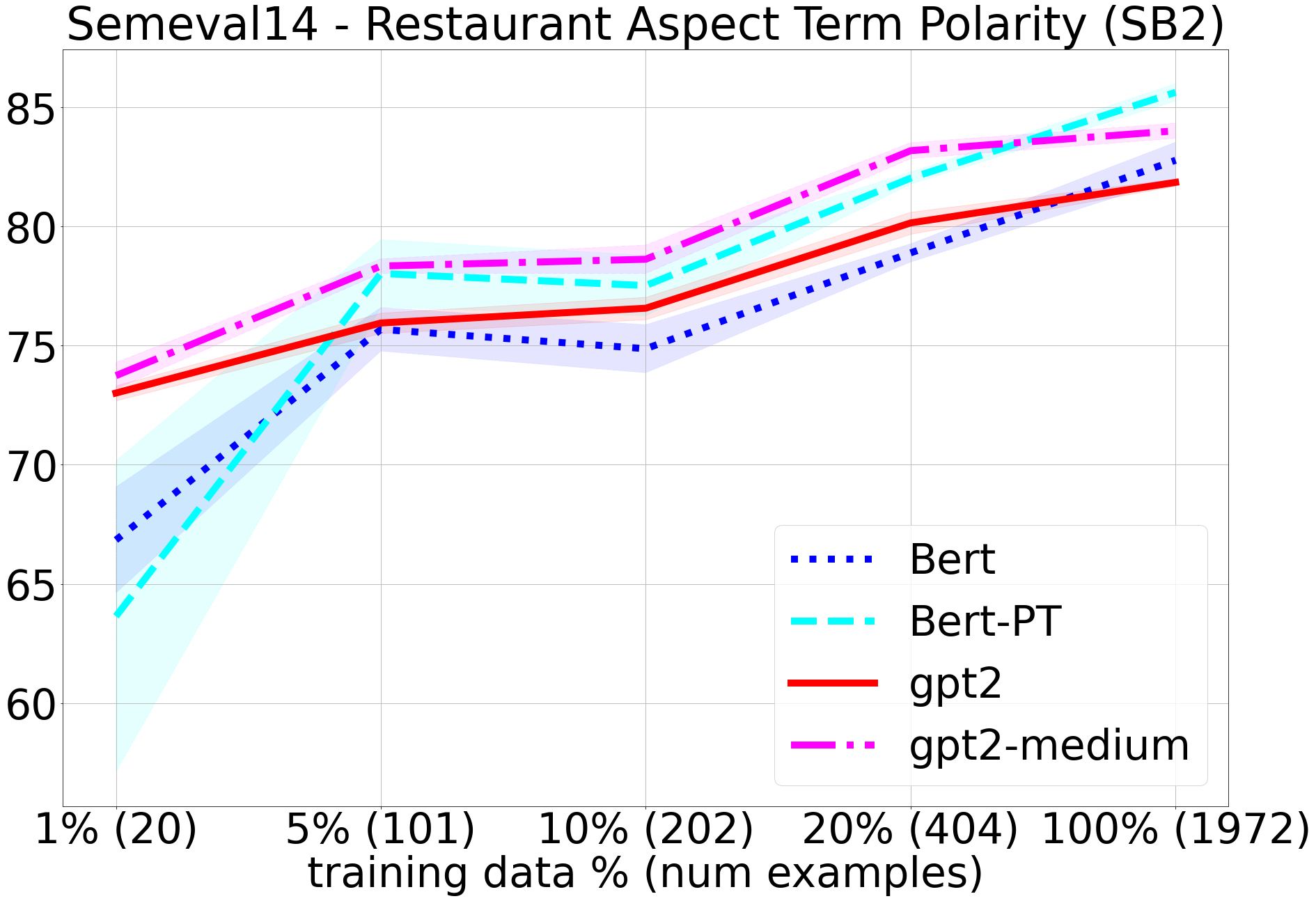}}
    \subfigure[
    % semeval 14, restaurants category polarity (SB4)
    ]{\includegraphics[width=0.3\linewidth]{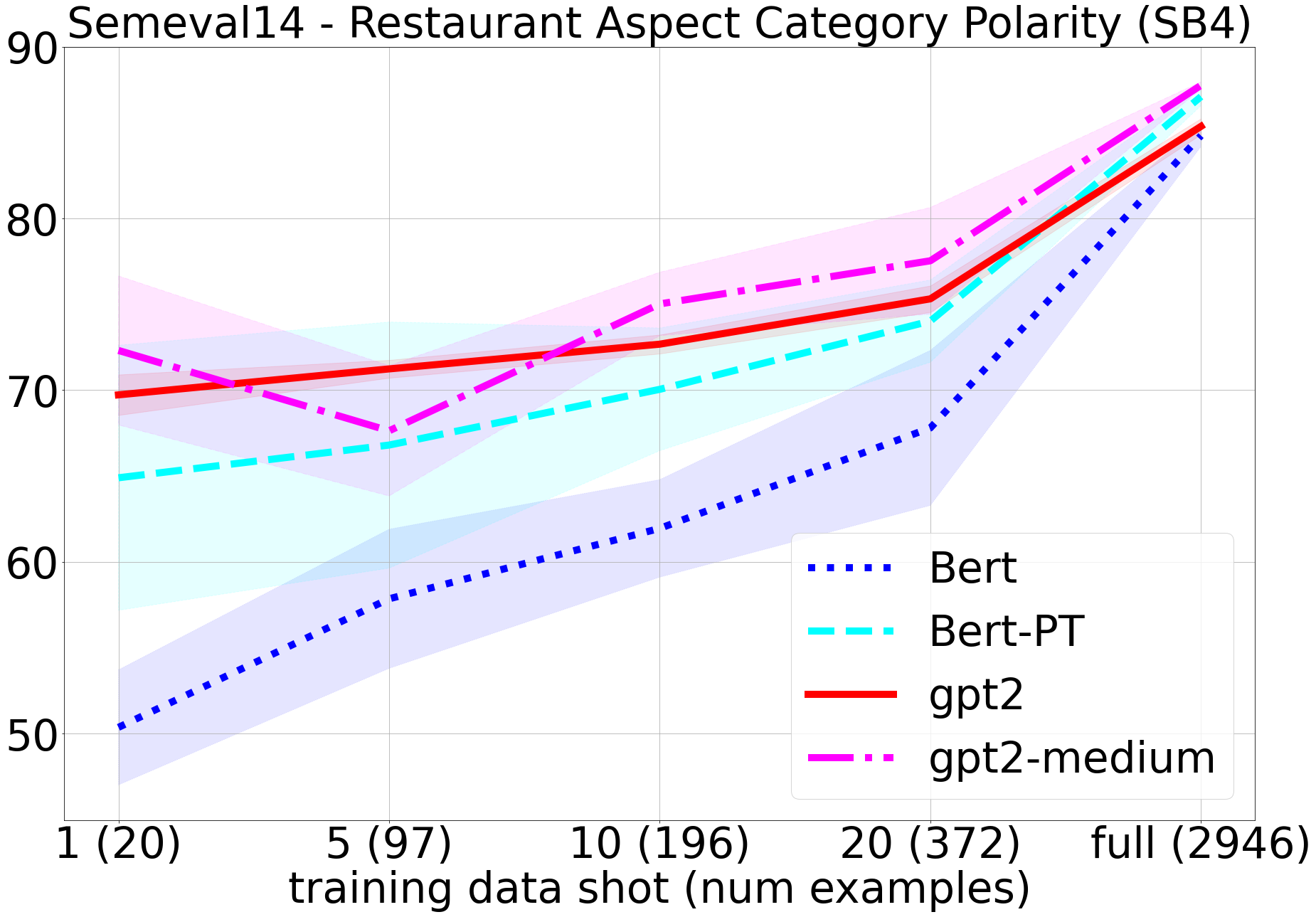}}
    \subfigure[
    % semeval 14, laptops term polarity (SB1)
    ]{\includegraphics[width=0.3\linewidth]{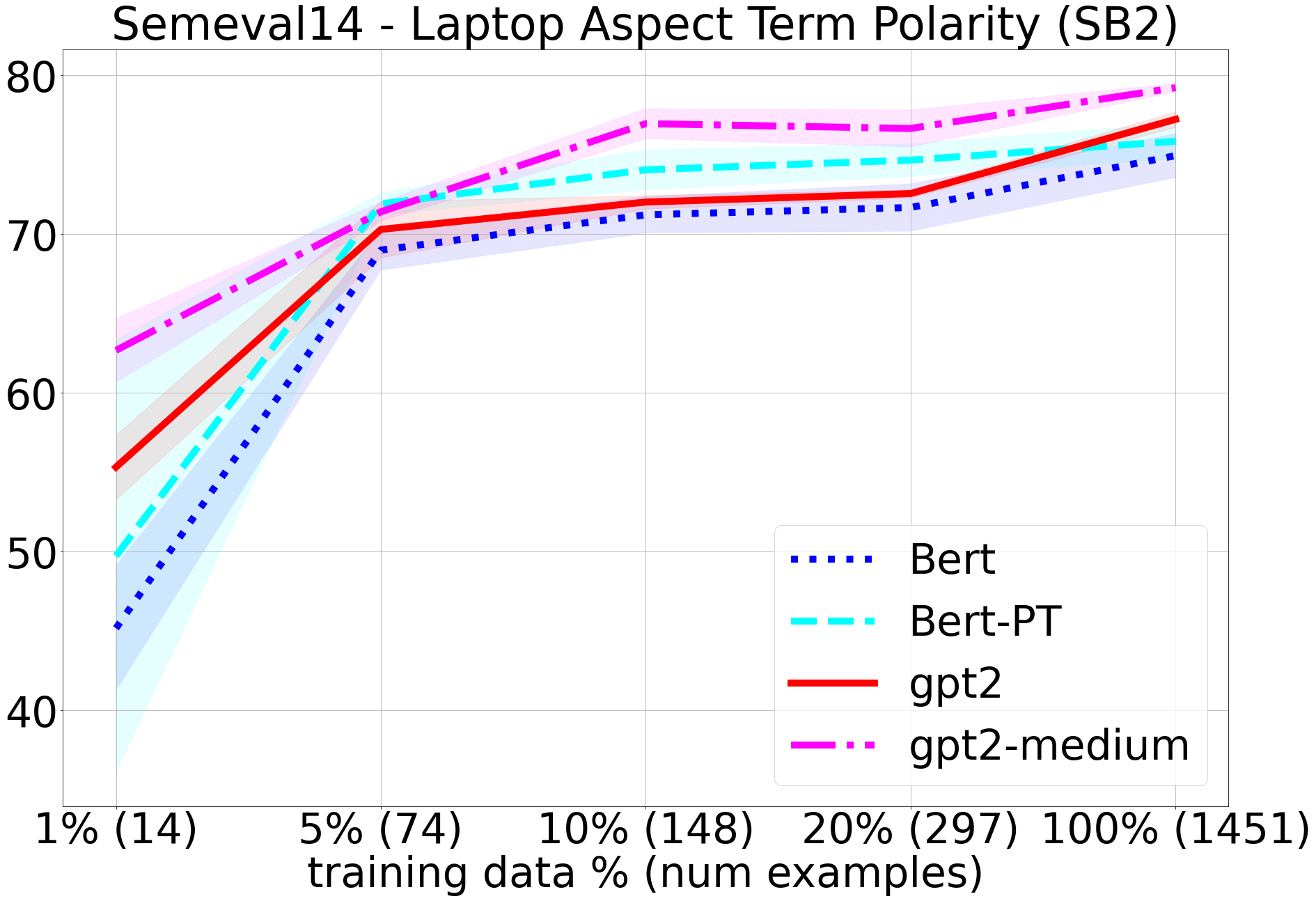}}

    \subfigure[
    % semeval16, restaurant term polarity (SB1)
    ]{\includegraphics[width=0.3\linewidth]{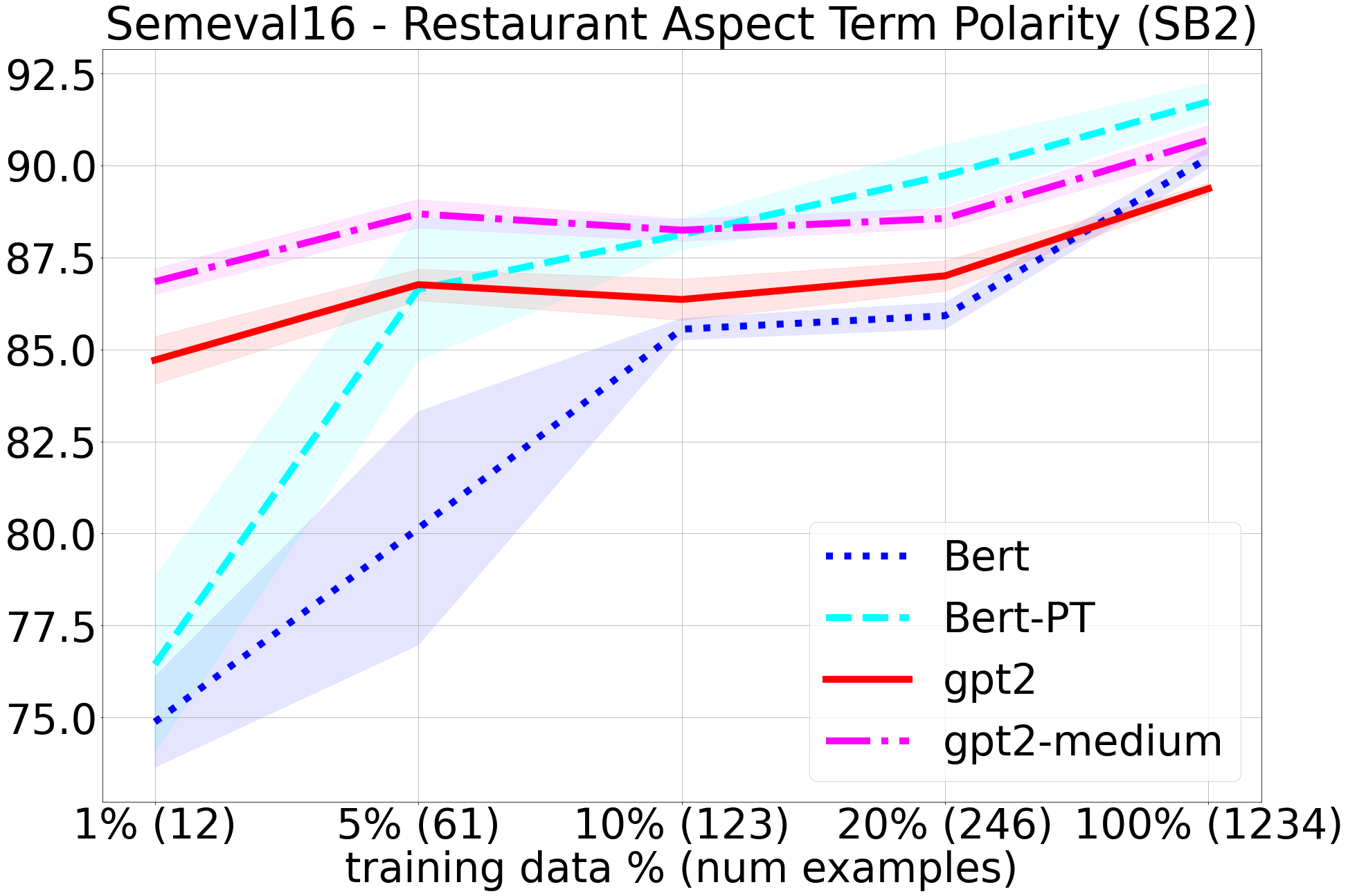}}
    \subfigure[
    % semeval16, restaurant category polarity (SB4)
    ]{\includegraphics[width=0.3\linewidth]{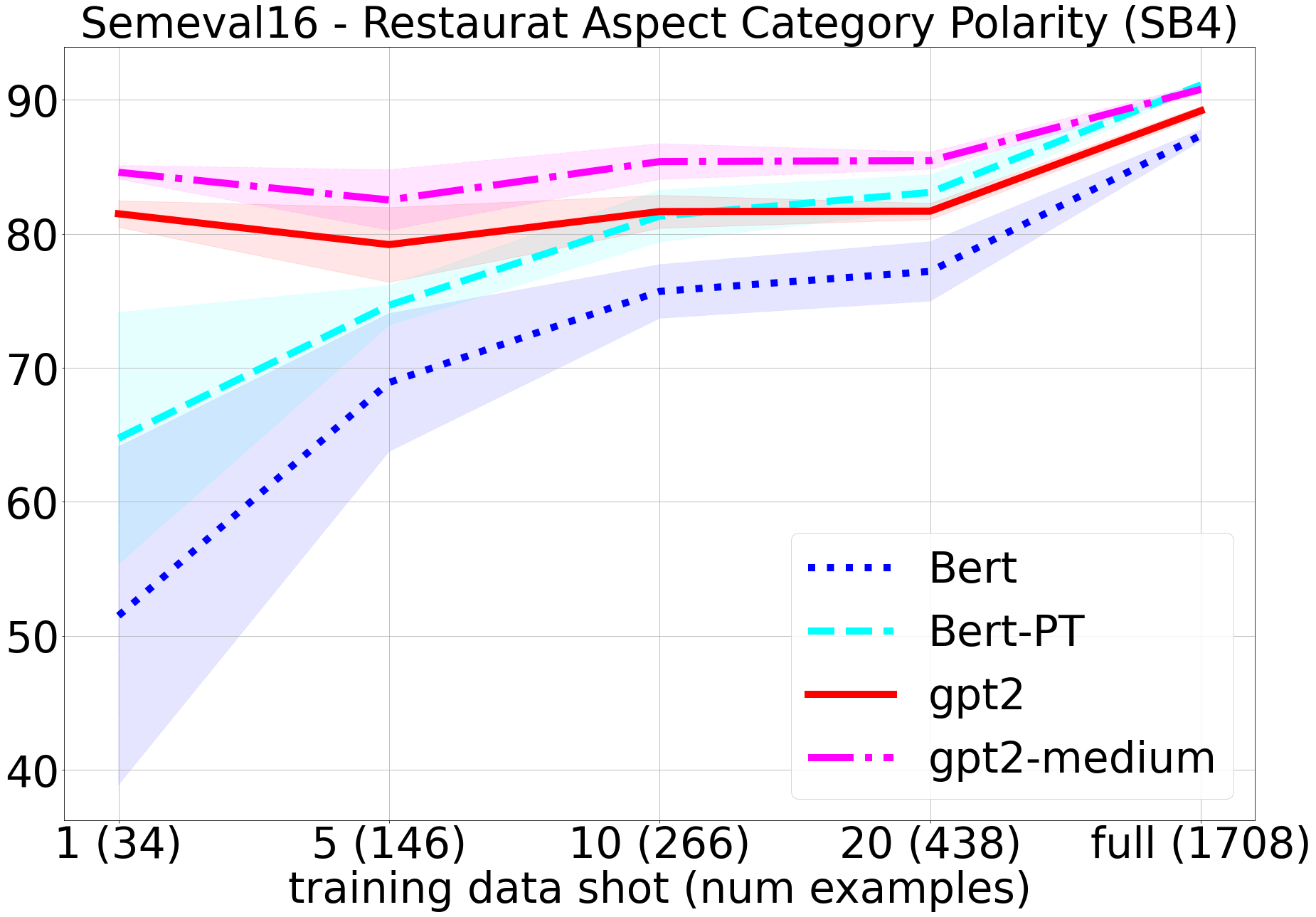}}
    \subfigure[
    % semeval 16, laptop category polarity (SB4)
    ]{\includegraphics[width=0.3\linewidth]{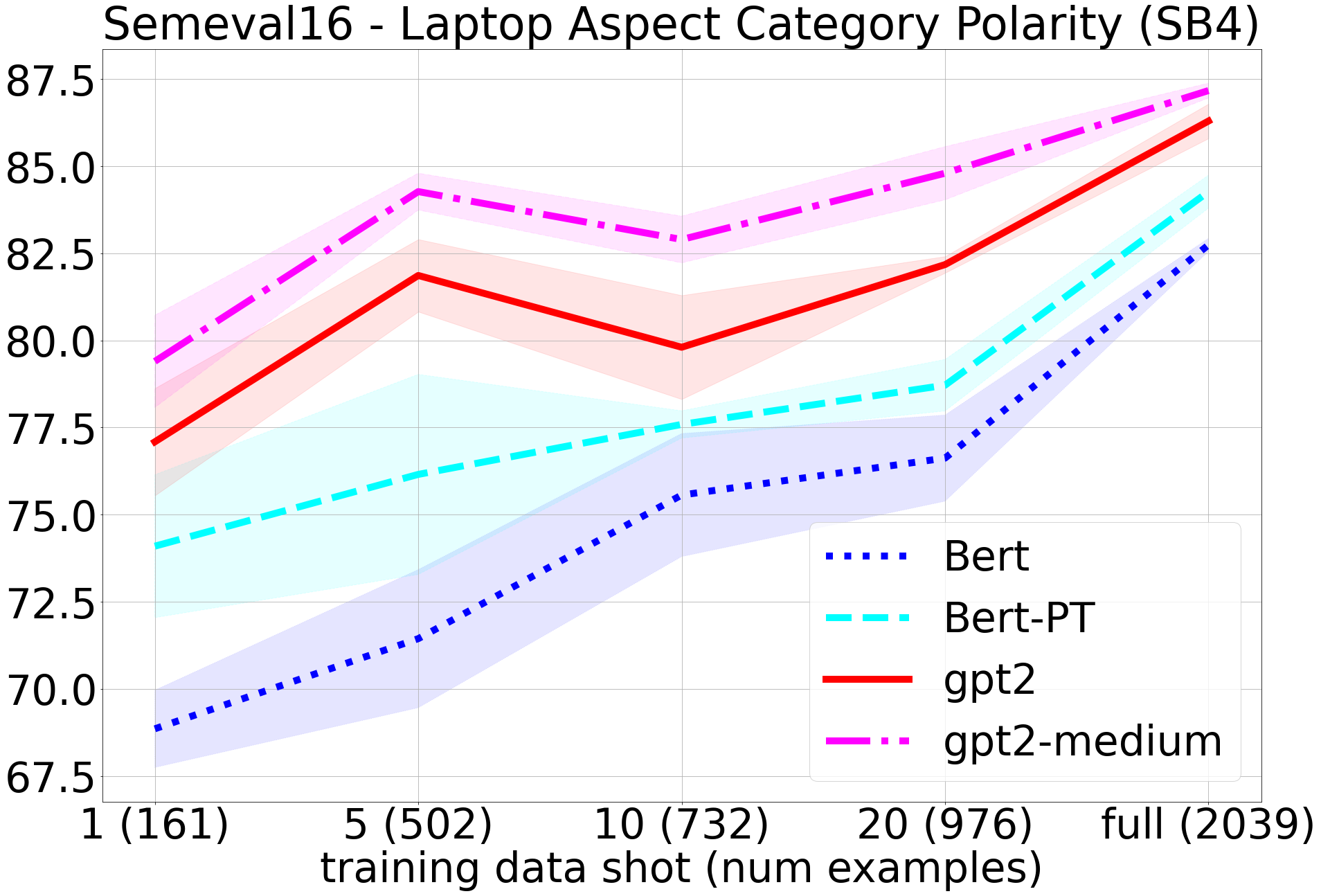}}
    \vspace{-3mm}
    \caption{Single-Task polarity prediction (SB2 and SB4 sub-tasks), in few and full-shot settings. Note: 1-shot refers to one example per class, for aspect category, and $1\%$ is percentage of training data for aspect term. Lines represents mean accuracy, and shaded area are standard deviation of experiments with 4 different random seeds.
    % For example, 1\% is about 20 examples for training.} 
    (best viewed in color)
    % \wl{we should avoid simply using the word ``shots" to express how many training samples are used for model training or the percentage of training samples used for model training. when we refer to a percentage of data used for training, we should also refer to the total number of training samples in the training set for clarity} \eh{see highlighted text}
    % \wl{why is there no \% in figure (d) and (f)?}
    % \eh{category datasets are based on shots, number of examples per class}
    } 
    \label{fig:single_task_polarity}
    % \end{subfigure}
    \vspace{-3mm}
\end{figure*}

\subsection{Dataset}
% In this paper, 
The proposed generative language model is evaluated on the two datasets proposed for ABSA task. SemEval14 challenge~\cite{pontiki-etal-2014-semeval} consists of four sub-tasks as described in section~\ref{subsec:absa_task}. We also evaluate the proposed model on task 5 of SemEval16~\cite{pontiki2016semeval}, which contains two sub-tasks for sentence and text level review data in multiple languages.
In this paper, we only focus on the English language of sub-task 1 (sentence level) to be able to compare with the prior arts. 

Moreover, we evaluate on Stanford Sentiment Treebank (SST) dataset~\cite{socher-etal-2013-recursive} for binary (SST-2) and fine-grained (SST-5) sentiment classification of movie reviews domain. Since intent detection is a similar task to sentiment analysis, the evaluation is also performed on out-of-scope (OOS) intent detection dataset~\cite{larson-etal-2019-evaluation} which created for chatbot systems.

To evaluate the performance
% proposed model and the BERT-based baselines 
on few-shot setting, 
we sub-sample training set for aspect term and aspect category domains. For aspect term, the train set is randomly sub-sampled to the smaller sizes, $[1\%, 5\%, 10\%, 20\%]$. For example, $1\%$ few-shot train set contains only about $\approx 20$ sentences.
% mention how many training example for few shot, for example 1%
For aspect category, since there is the predefined set of categories, we randomly sub-sample examples for each category, with different number of examples of $[1, 5, 10, 20]$.

The distribution of the train, dev and test splits for each domain are shown in Table~\ref{tab:dataset}. It is noteworthy that the previous baselines have created customized validation set from train set. Since no official validation set is released for SemEval14 and SemEval16, and in order to have a unified evaluation, we used the official \verb|trial| set (part of train set) for validation,
% set, 
and exclude those examples from the train set. Moreover, 
% previous state-of-the-arts 
prior works
excluded examples with \verb|conflict| polarity from their evaluations, since it is considered a difficult prediction task. However, for more accurate evaluation,
% to analyze the ability of our proposed model, 
these examples are retained in our evaluation.

\vspace{-2mm}
\subsection{Evaluation}
Performance evaluation of aspect term polarity (SB2) and aspect category polarity (SB4) single-tasks in Eq.~(\ref{eq:single-task_term}) and Eq.~(\ref{eq:single-task_category}) are based on accuracy metric. It is measured by counting the number of aspect term and aspect category polarities which are correctly predicted. The evaluation of aspect term extraction (SB1) and aspect category detection (SB3) are measured by F1 metric~\cite{pontiki-etal-2014-semeval} computed on the overlap of the ground-truth and generated sequences. The evaluation of SST-2, SST-5 and OOS datasets are measured by accuracy metric. On OOS dataset, full accuracy on in-domain and out-of-scope examples are measured.

Evaluation of joint and multi-task models in Eq.~(\ref{eq:joint-task_term})(\ref{eq:joint-task_category})(\ref{eq:multi-task}) are measured by joint accuracy. This means that for an example sentence $S^{k}$, if all the aspect term and term polarity predictions are correct, it is assumed as a correct prediction. 
% for~Eq.~(\ref{eq:joint-task_term}).

% For laptop domain, it is annotated with aspect term only in SemEval14, and with aspect category only in SemEval16. Therefore, evaluation on this domain are constrained to the annotation.

The restaurant domain contains both aspect term and aspect category annotations for SemEval14 and SemEval16. However, the laptop domain only contains aspect term annotation for SemEval14, and aspect category annotation for SemEval16. Therefore, single-task evaluation on laptop domain is constrained and multi-task prediction performance can only be evaluated on restaurant domain.

\vspace{-2mm}
\section{Experiments}
\label{sec:exp}
% \vspace{-2mm}

\begin{table*}[htb!]
    \centering
    % \small
    \scriptsize
    % \begin{tabular}{ccccccc}
    \begin{tabular}{lllrrrr}
    \hline
        \multirow{2}{*}{Method} & \multirow{2}{*}{Training Task} & \multirow{2}{*}{Model} & \multicolumn{2}{c}{Restaurant} & \multicolumn{2}{c}{Laptop}   \\
        
        & & & Joint Accuracy & SB1 (F1) & Joint Accuracy & SB1 (F1) \\
        \hline
        % \cline{3-7}
        \multirow{7}{*}{Discriminative} & \multirow{7}{*}{Single (SB1)} & MGAN & - &  $71.48$ & - & $71.42$\\
        & & BERT & - & $74.1$ & - & $79.28$ \\
        %  & BERT & - & & - & \\
         & & BERT-DK & - & $77.02$ & - & $83.55$ \\
         & & BERT-MRC & - & $74.21$ & - & $81.06$ \\
         & & BERT-PT & - & $77.97$ & - & $84.26$ \\
         & & BERT-PSUM & - & - & - & $85.94$\\
         & & BERT-HSUM & - & - & - & $\boldsymbol{86.09}$ \\
         \hline 
         \multirow{4}{*}{Generative} & \multirow{2}{*}{Joint (SB1\&2) } & GPT2 (base) & 
        %  54.79 & 76.61  & 49.05 & 71.42 \\
         $56.47_{\pm 0.82}$ & $77.59_{\pm 0.32}$ & $50.65_{\pm 1.04}$ & $72.61_{\pm 1.03}$ \\
         & & GPT2 (medium) &  
        %  59.9 & \textbf{80.23} &  53.79 & 74.33 \\
        $60.07_{\pm 0.52}$ & $\boldsymbol{81.52}_{\pm 0.8}$ & $53.55_{\pm 0.43}$ & $75.94_{\pm 0.17}$ \\
         \cline{2-7}
         & \multirow{2}{*}{Multi (SB1-4)} & GPT2 (base) & %51.63 & 77.43  & - & - \\
         $49.84_{\pm 1.03}$ & $77.92_{\pm 0.53}$ & - & - \\
         & & GPT2 (medium) 
        %  & 55.62 &  \textbf{81.53} & - & - \\
        & $54.43_{\pm 0.47}$ & $\boldsymbol{82.04}_{\pm 0.21}$ & - & - \\
         \hline

    \end{tabular}
    \caption{SemEval14 SB1 and SB2 sub-tasks for restaurant and laptop domains. Comparing joint and multi-task generative model with single-task BERT baselines for full-shot setting.}
    \label{tab:joint_task}
    \vspace{-3mm}
\end{table*}

% In this section, the evaluation results are presented.
% on single, joint and multi-task predictions are presented for few- and full-shot settings. 
The proposed generative language model is evaluated on five tasks. Single-task setting includes aspect term polarity and aspect category polarity prediction, Eq.~(\ref{eq:single-task_term})(\ref{eq:single-task_category}), for restaurant and laptop domains. Joint-task includes a) aspect term extraction and polarity~Eq.~(\ref{eq:joint-task_term}) and b) aspect category detection and polarity~Eq.~(\ref{eq:joint-task_category}). Finally, multi-task setting comprises all sub-tasks, i.e. aspect term extraction (SB1), aspect category detection (SB3), and their polarity predictions (SB2 and SB4), Eq.~(\ref{eq:multi-task}).
% , for restaurant and laptop domains. 

The evaluation of our proposed generative language model is compared with recent 
% BERT-based model, 
BERT-PT~\cite{xu2019bert} model. We have reproduced results of BERT-PT on full-shot settings, since we include examples with \verb|conflict| polarity. Other BERT-based models such as BERT-IL~\cite{reddy2020does} has not open-sourced code, and therefore they are not included in few-shot evaluation.

\vspace{-2mm}
\subsection{Single-Task Polarity evaluation}
\label{exp:single_task}

In this section, the proposed generative language model is evaluated on aspect term and aspect category polarity prediction for both restaurant and laptop domains. As shown in Figure~\ref{fig:single_task_polarity}, the proposed model, based on \textit{GPT2-base}, outperforms BERT on few- and full-shot settings on all sub-tasks (SB2 and SB4) for SemEval14 and SemEval16. More importantly, GPT2 model has lower variance than BERT, especially in $1\%$ or 1-shot setting.

It is shown that BERT average performance drops by a large margin on low-resource regimes ($<5\%$ or $<5$ shot) and with increased variance, whereas our proposed generative model shows robust performance on few-shot setting with small variance. Compared to BERT-PT~\cite{xu2019bert}, which exploits additional pre-training on review data from Amazon and Yelp datasets, and using auxiliary tasks of MRC, generative model with more layers (\textit{GPT2-medium}) and no additional pretraining matches or outperforms BERT-PT average performance
% its average performance or outperforms BERT-PT 
in few-shot setting with smaller variance.
% , without any additional pre-training. 
Interestingly, \textit{GPT2-base} model (12 layers) 
% even 
outperforms BERT-PT average performance in some cases, including all $1\%$ and $1$-shot settings with reduced variance. For example, \textit{GPT2-base} 
% performance
outperforms by a large margin, $16.75$ points on average accuracy and reduces standard deviation by $8.8$ points on $1\%$-shot setting of category polarity prediction in restaurant domain of SemEval16, Figure~\ref{fig:single_task_polarity}(e). 
% , except in~Figure~\ref{fig:single_task_polarity}(c) for laptop domain. 
Moreover, \textit{GPT2-base} outperforms BERT-PT in all few- and full-shot settings on aspect category polarity prediction task (SB4) of restaurant domains in SemEval16 dataset, ~Figure~\ref{fig:single_task_polarity}(f). 

Although \textit{GPT2-medium} average performance mostly outperforms BERT-PT, there are some exceptions, such as Figure~\ref{fig:single_task_polarity}(a) for full-shot, Figure~\ref{fig:single_task_polarity}(c) for $5\%$-shot, Figure~\ref{fig:single_task_polarity}(d) for $20\%$ and full-shot. On the other hand, BERT-PT has much larger variance and less robustness in all few- and full-shot settings.
% ~Figure~\ref{fig:single_task_polarity}(b) for $5\%$ and $20\%$, and~Figure~\ref{fig:single_task_polarity}(c) for $5\%$, and~Figure~\ref{fig:single_task_polarity}(e) for $10\%$. 
% Moreover, BERT-PT also outperforms \textit{GPT2 medium} on full-shot setting on three sub-tasks, as shown in~Figure~\ref{fig:single_task_polarity}(c)(e)(f). 
This is perhaps due to the use of out-of-domain data in additional pre-training of BERT-PT which results in higher variance, even than BERT baseline, when finetuned on few-shot downstream tasks. 
% This might also overlaps with the fine-tuning task, especially for some of the few-shot splits, which improves the model performance on fine-tuned task. 
The goal of our proposed model is not to simply outperforms BERT-PT by additional pre-training, but to provide a robust model for few-shot setting.

\begin{figure}[htb!]
    \centering
    \includegraphics[width=0.8\linewidth]{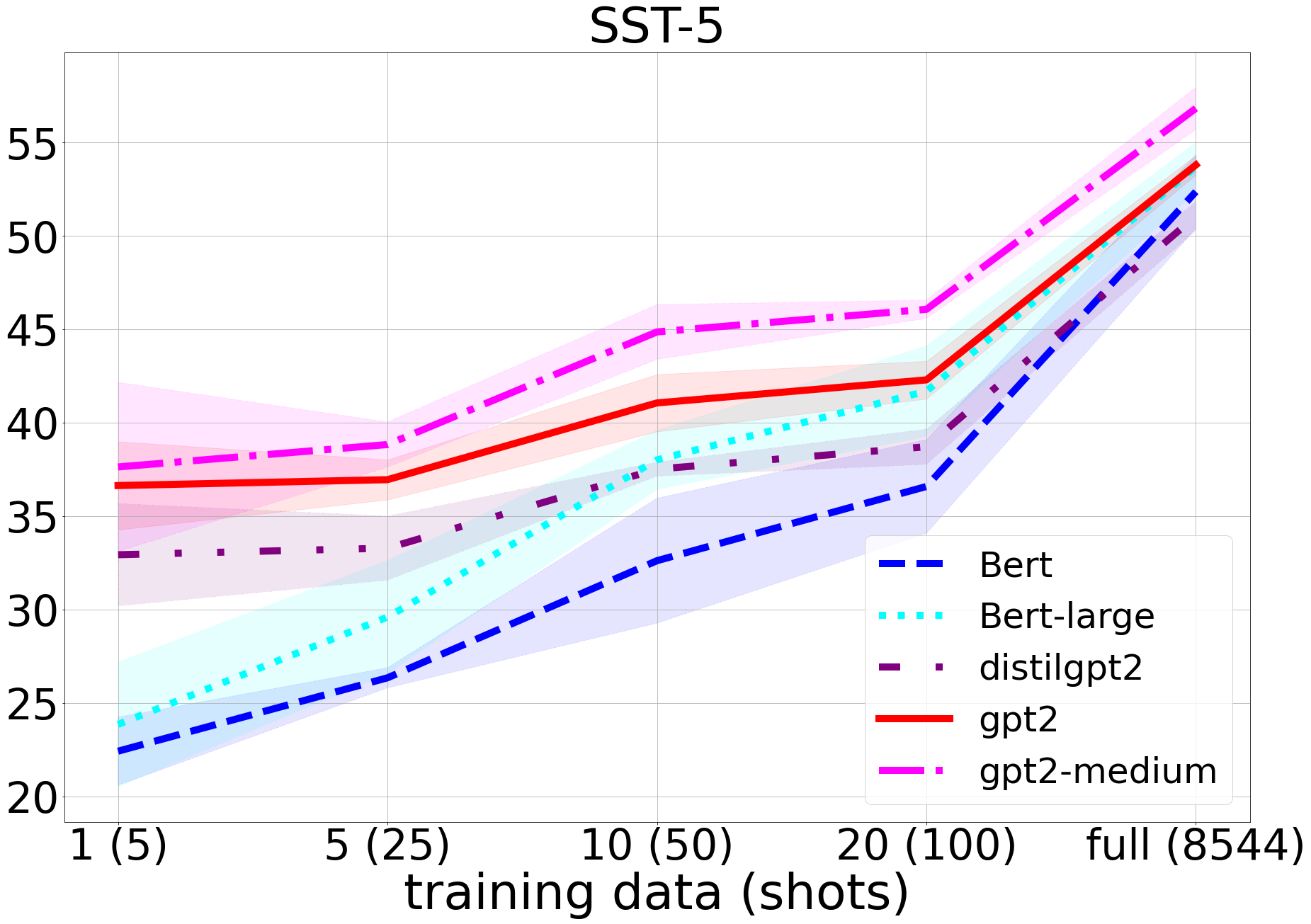}
    \caption{
    % Few-shot evaluation 
    % on SST2 dev set. 
    Few-shot evaluation 
    on SST5 dataset. 
    Note: 1-shot refers to one example per class. 
    (best viewed in color)}
    \label{fig:sst5}
    \vspace{-3mm}
\end{figure}

More evaluation on sentiment polarity prediction on SST5, SST2 and OOS intent detection datasets are presented in Figure~\ref{fig:sst5}, Appendix~\ref{sec:appendix_other_sentiment} and Figure~\ref{fig:ablation_other_sentiment_appendix}. They indicate that generative language model outperforms BERT-based classifier models. Overall, the results of single-task polarity prediction indicate that our proposed generative model based on language generation (uni-directional self-attention) have better performance than the discriminative models which uses BERT (bi-directional self-attention) as encoder.

\vspace{-2mm}
\subsection{Joint and Multi-Task evaluation}
\label{exp:joint_task}
In this section, the proposed generative model is evaluated for joint and multi-task prediction. It includes solving two sub-tasks jointly, e.g. aspect term extraction and term polarity prediction, or aspect category detection and category polarity prediction, Eqs.~(\ref{eq:joint-task_term})(\ref{eq:joint-task_category}), or predicting all Eqs.~(\ref{eq:multi-task}).
Since BERT and BERT-PT are single-task models, which required to use different heads for each sub-task, we can not directly compare our joint-task model with these baselines on joint-accuracy metric. For example, BERT-PT uses groundtruth aspect term to evaluate on polarity prediction (SB2), which is not comparable to our joint-task model which generates aspect term and polarity jointly.

Results in Table~\ref{tab:joint_task} indicate that although generative model is trained in joint-task manner, for predicting aspect term extraction and term polarity, it still outperforms BERT-PT and other BERT baselines which are trained to solve single-task aspect term extraction only, on aspect term extraction (SB1) metric, in restaurant domain. However, in laptop domain, the generative model under-performs BERT-based models on aspect term extraction (SB1) metric, perhaps due to less training data in laptop domain for joint-task loss.

% \begin{figure*}[htb!]
%     \centering
%     \subfigure[
%     % semval14, restaurant term polarity
%     ]{\includegraphics[width=0.3\linewidth]{figures/ablation_semeval14_res_term_polarity.png}}
%     \subfigure[
%     % semval14, laptop term polarity
%     ]{\includegraphics[width=0.3\linewidth]{figures/ablation_semeval14_laptop_term_polarity.png}}
%     \subfigure[
%     % semval16, restaurant term polarity
%     ]{\includegraphics[width=0.3\linewidth]{figures/ablation_semeval16_res_sb1_term_polarity.png}}
%     % \vspace{-2mm}
%     \caption{Ablation analysis on model input sequence formatting. \textit{GPT2 (split)} means review sentence is concatenated with each aspect terms separately. (best viewed in color)}
%     \label{fig:ablation}
%     % \vspace{-2mm}
% \end{figure*}

% \subsection{Multi-Task prediction}
% \label{exp:multi_task_learning}
% The proposed generative language model is inherently a multi-task model, which can solve all sub-tasks (SB1-4) as language generation. During training, a single review sentence is concatenated with all aspect terms and categories and their polarities. 

% The restaurant domain contains both aspect term and aspect category for SemEval14 and SemEval16. However, the laptop domain contains only aspect term annotation for SemEval14, and only aspect category annotation for SemEval16. Therefore, multi-task training can only be evaluated on restaurant domain. 
\vspace{-3mm}
\paragraph{Aspect category sub-tasks improve aspect term extraction:} In multi-task setting, where generative model is trained on all sub-tasks (SB1-4), the aspect term extraction (SB1) F1 metric is improved more, compared to when trained as a single-task model. This indicates that training the generative model using extra supervision (from aspect category) helps to extract multiple aspect terms in the review sentence more accurately.

\vspace{-2mm}
\paragraph{Generative language modeling is better for multi-task learning:} Evaluation results on SemEval14 restaurant domain are shown in Appendix~\ref{sec:appendix_multitasking} Table~\ref{tab:multi_task_semeval14_restaurant}. Combined with the results from Table~\ref{tab:joint_task}, it indicates that the proposed generative language model performs well on solving all sub-tasks (SB1-4) using language generation. 
% , and no task-specific layer. 
For example, compared to joint-task setting (Table~\ref{tab:joint_task}), aspect term extraction (SB1) F1 metric improves more for restaurant domain. Multi-task evaluation results on SemEval16 restaurant domain are shown in Appendix~\ref{sec:appendix_multitasking} Table~\ref{tab:multi_task_semeval16_restaurant} for reference.

\begin{figure*}[htb!]
    \centering
    \subfigure[
    ]{\includegraphics[width=0.3\linewidth]{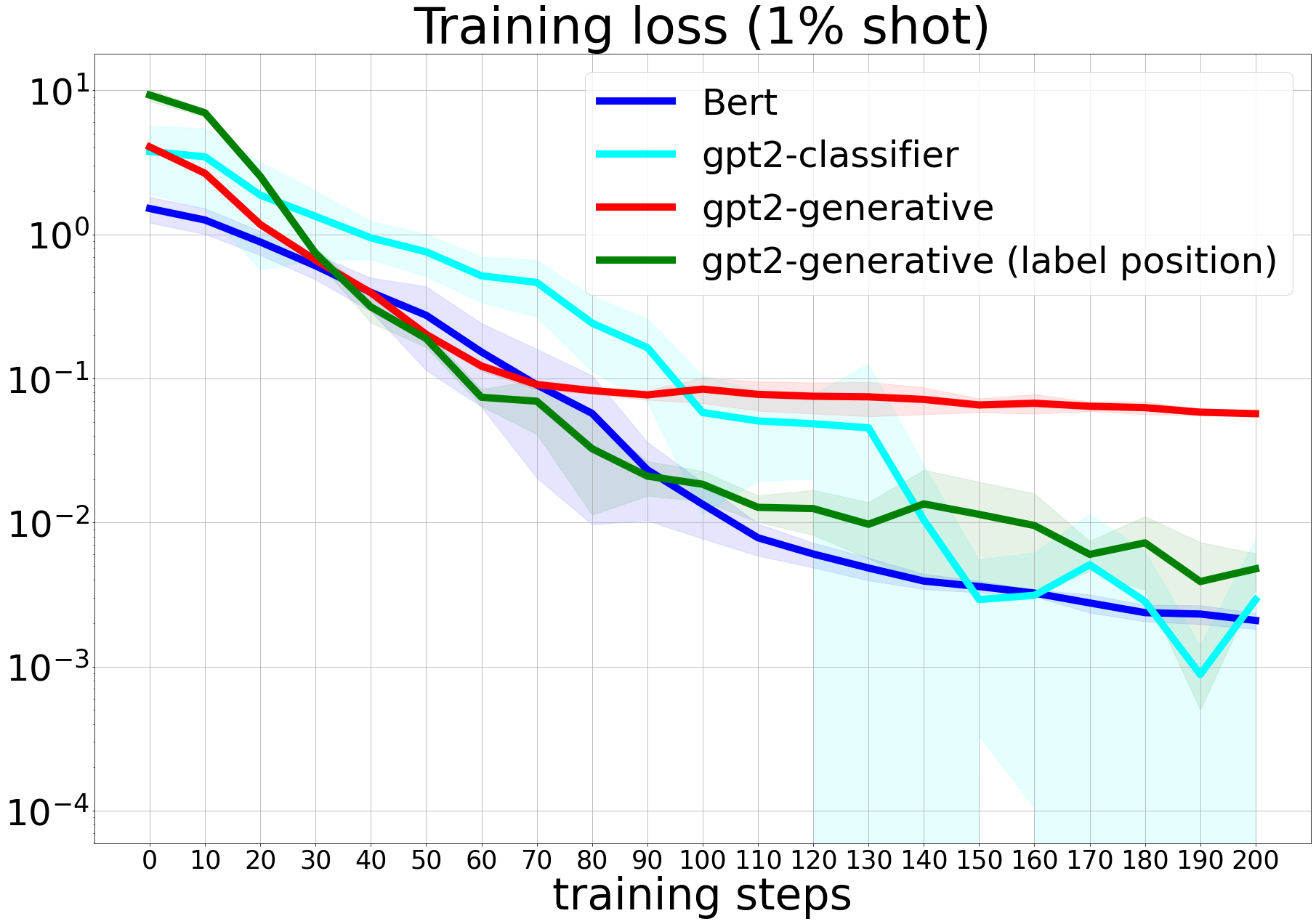}\label{subfig1}}
    \subfigure[
    ]{\includegraphics[width=0.29\linewidth]{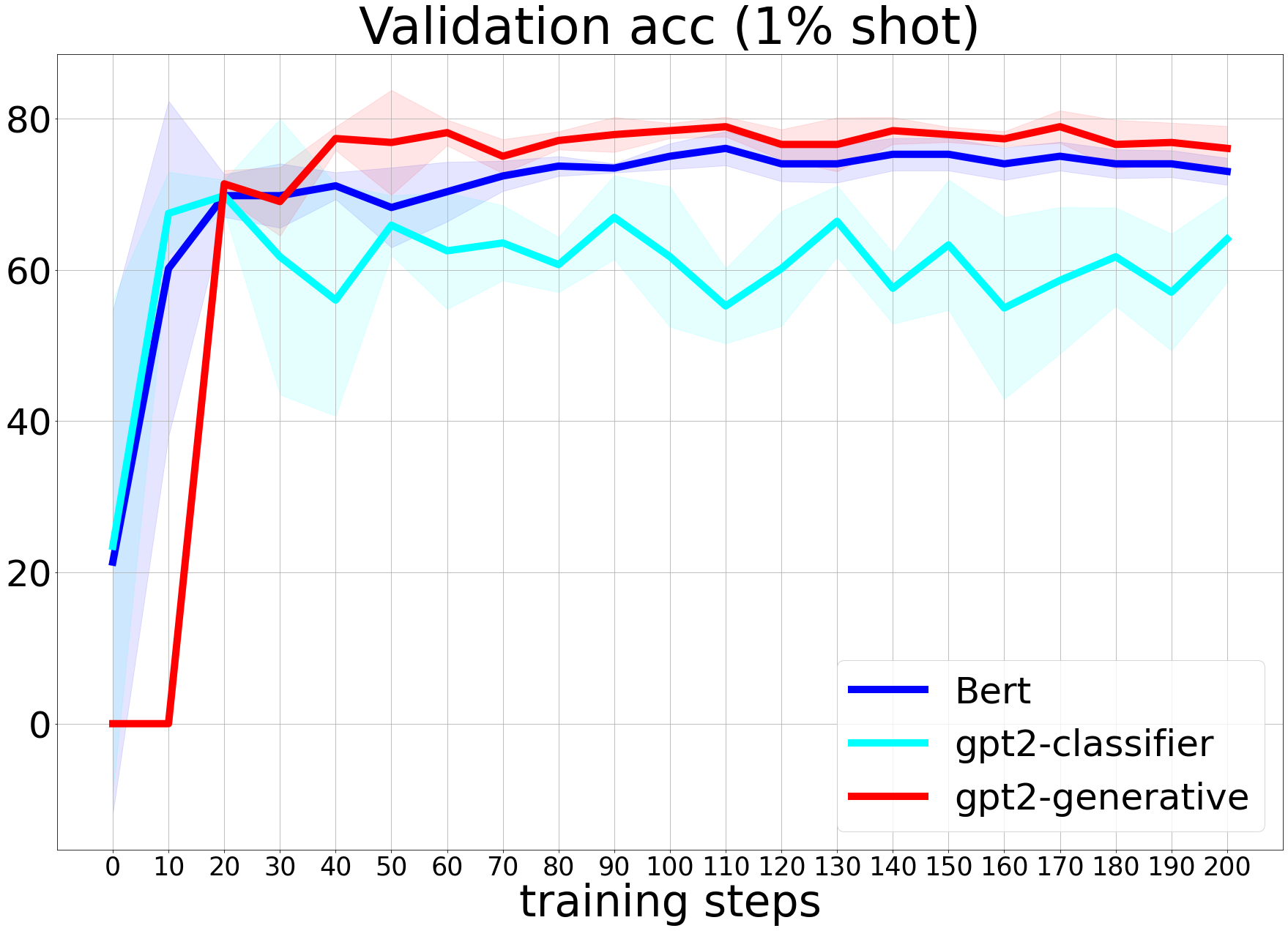}\label{subfig2}}
    \subfigure[
    ]{\includegraphics[width=0.31\linewidth]{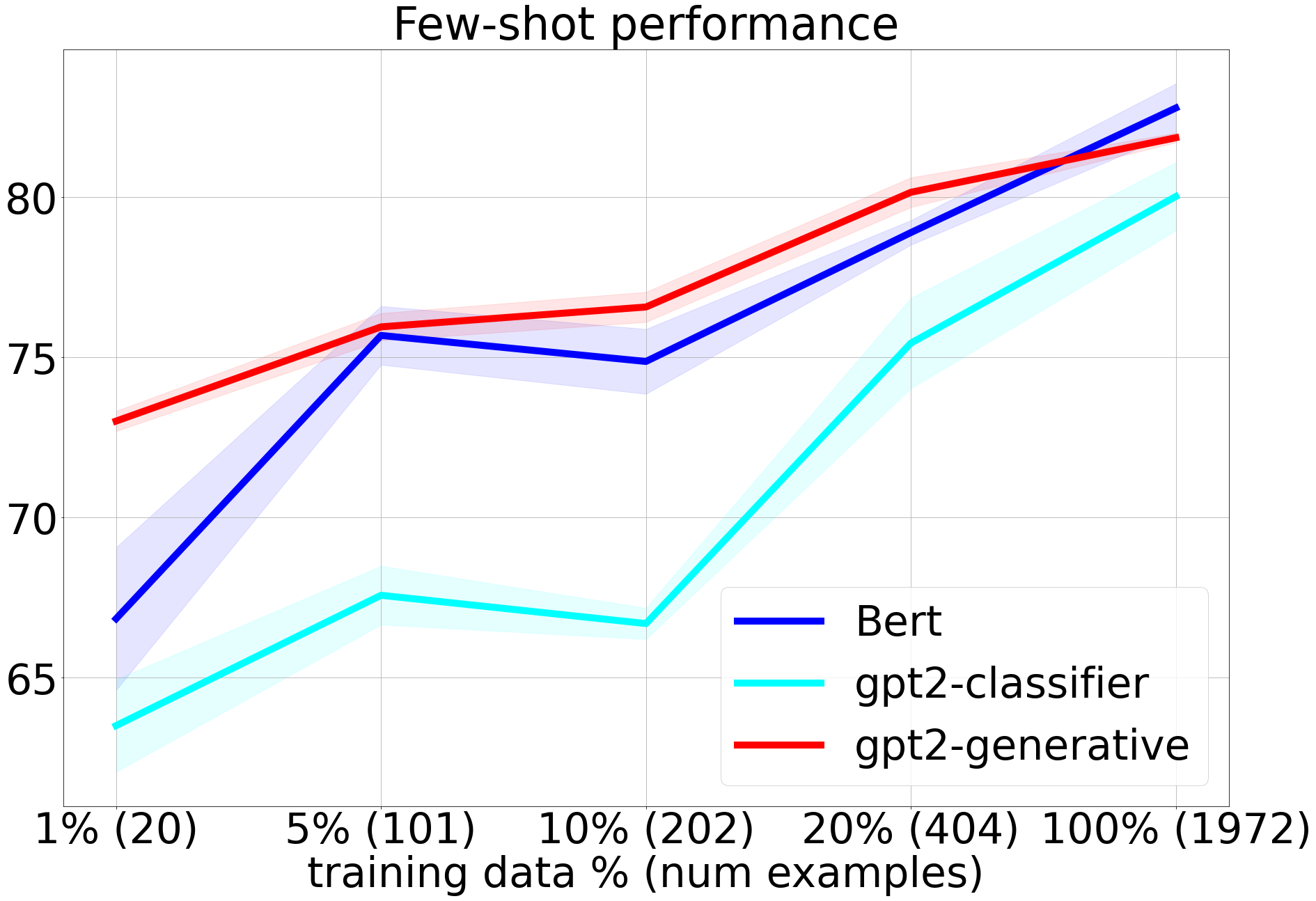}}
    % \subfigure[SemEval14 restaurant aspect term polarity (SB2)
    % ]{\includegraphics[width=0.7\linewidth]{figures/ablation/training_convergence/semeval14_restaurant_term_train_val.png}\label{subfig2}}
    
    % \subfigure[
    % ]{\includegraphics[width=0.3\linewidth]{figures/bert_gpt2_test_acc.png}}
    \vspace{-3mm}
    \caption{Analysis of few-shot training convergence, evaluated on SemEval14 aspect term polarity prediction (SB2) on restaurant domain for $1\%$ training data. GPT2-classifier model uses a classification layer on the output of last input token without using language modeling loss for training. \textit{Note: Lines represents mean value, and shaded area are standard deviation of experiments with 4 random seeds.}
    % GPT2 language modeling loss converges slower than BERT. However, GPT2 loss on position for predicting label converges faster than BERT, results in better performance between 40-90 steps. 
    % \wl{IF you believe this figure helps shed lights on the superiority of GPT2 model in few-shot settings, I recommend running the same analysis for more tasks and dataset - hanging the insight solely on one dataset and one task appears thin. Also, I am not sure it's the most rigorous thing to do to plot test performance curve} \eh{will investigate on other datasets} \eh{gpt2-classifier is added}
    (best viewed in color)}
    \label{fig:ablation_convergence}
    \vspace{-4mm}
\end{figure*}

% \begin{figure*}[htb!]
%     \centering
%     \subfigure[
%     ]{\includegraphics[width=0.3\linewidth]{figures/sst2.png}}
%     \subfigure[
%     ]{\includegraphics[width=0.3\linewidth]{figures/sst5.png}}
%     \subfigure[
%     ]{\includegraphics[width=0.3\linewidth]{figures/intent_acc_full.png}}
%     % \vspace{-2mm}
%     \caption{Few-shot evaluation of GPT2 and BERT models on SST2 dev set, SST5 and OOS intent detection datasets. Note: 1-shot refers to one example per class. For aspect term, the percentage of training data are used for few shot settings. 
%     % For example, 1\% is about 20 examples for training.} 
%     (best viewed in color) 
%     % \wl{see comments before, let's avoid using ``shots" and explain what it actually represent. ALso, what are the considerations going into these comparisons? Why was distilgpt2 added? it is not competitive in all settings. How many runs are there for these plots? the performances seems to be not very sable and there are strange bumps and drops.} \eh{The purpose is to show the performance of generative model with different number of parameters with BERT. These are 4 different runs for intent detection. The purpose of distilgpt2 is to show its performance equality to Bert-base model } \wl{is this Figure referred to anywhere in the paper?} \eh{check paragraph before section 4.5}
%     }
%     \label{fig:ablation_other_sentiment}
%     % \vspace{-2mm}
% \end{figure*}

\vspace{-2mm}
\subsection{Ablation}
\label{exp:ablation}
% \vspace{-1mm}
In this section, 
% we present 
the ablation study of proposed generative language model is studied on two aspects. First, 
% investigating 
using the language model (GPT2) as a discriminative classifier vs. for language generation. Second, we study the training convergence of generative model with two discriminative baselines, i.e. BERT and GPT2 as classifier to better understand few-shot performance.

\paragraph{Generative vs. Discriminative training of unidirectional language model:} To analyze the benefit of fine-tuning GPT2 using language modeling loss, we also fine-tune it as a classifier. In the latter case, a classification layer is added, which uses the output of the last token of the input sequence for polarity prediction.
% (SB2 and SB4). 
As shown in 
% Figure~\ref{fig:ablation_generative_restaurant}, 
Figure~\ref{fig:ablation_convergence}(c),
\textit{GPT2-classifier} under-performs BERT, when only trained with discriminative loss. We conjecture that since GPT2 uses uni-directional self-attention (left-to-right), it captures less contextualized representation, compared to bidirectional self-attention in BERT. 
On the other hand, 
% In contrast, 
when fine-tuning GPT2 using generative loss (next word prediction), uni-directional self-attention learns a better representation, which improves few-shot performance. Ablation analysis on laptop domain and aspect category polarity predictions for both domains are shown in Appendix~\ref{sec:appendix_ablation_generative} and 
Figures~\ref{fig:appendix_ablation_convergence_semeval14} and~\ref{fig:appendix_ablation_convergence_semeval16}.
% Figure~\ref{fig:ablation_generative}. 

\vspace{-3mm}
\paragraph{GPT2 language model exploits more supervision than BERT in few-shot setting:} To understand the training dynamics of generative language model and its relation to few-shot performance, we investigate the training convergence for GPT2, BERT, and \textit{GPT2-classifier}. 
Results for SemEval14 restaurant aspect term polarity prediction are shown in Figure~\ref{fig:ablation_convergence}. It is indicated that BERT model converges faster than GPT2 in 1\% few-shot settings, due to using a small classification head (fully-connected layer with 4 outputs) for the downstream task, which perhaps makes the model to overfits quickly to few-shot training data.
% (GPT2 uses output layer with 50k outputs)} 
% \wl{true but the entire BERT model, including all the transformer layers were also fine-tuned}. \wl{on 10/22 1) please address my previous comment 2) I recall that you reported performances of the model at thousands of steps but here you are suggesting that the convergence is reached before 200 hundred steps?} \eh{1) I meant that BERT is using a small fc layer for classification, versus GPT2 is using a large fc layer (size of vocabulary) for classification 2) in previous figures, every 400 steps was reported, and both models was still converging. we decided to do a new plot with 10-step validation up to 200 steps for better understanding} 
On the other hand, GPT2 converges more slowly, perhaps due to using language modeling loss, i.e. cross-entropy loss across all tokens of the input sequence, and also using output layer with size of the vocabulary. 
% for the downstream classification task. 
However, the cross-entropy loss on the position corresponding to predicting label,  \textit{gpt2-generative (label position)}, converges faster than BERT, early in training, and the loss value is smaller than BERT between 40-90 steps,  
% \wl{we should not be comparing the absolute values between two different models' training losses} \eh{?? this is comparing cross-entropy loss for predicting the same set of classes, between BERT and GPT2 (label position) curves}\wl{What matters is convergence. For example, the fact that the GPT2-label position loss covnerges to a higher loss value than BERT might not have any bearing on its performance. And I don't recommend plotting test set performance}. 
% \eh{here, I'm trying to day that few-shot performance is not related to convergence for downstream task. BERT converges to a lower value because of few training samples, and a loss which is specifically designed to converges to predicting correct class label. However, GPT2 loss is different and not intended to converges the downstream classification. it used new word prediction loss for all positions, which is the source of extra supervision. It also causes the loss of label position to be higher than BERT loss} 
% \wl{can we draw the causal connection here already? Later you said that the lower loss value of BERT didn't translate into better performance, here the writing is self-contradicting. If something is a hypothesis or a conjecture or educated guess, you should explicitly call it out, instead of using languages that suggest a more definitive conclusion than we have evidence for.} \eh{please read my previous comment related to this} 
where the model has better validation accuracy than BERT. Later during the training, BERT training loss converges to smaller values, but its performance does not outperform GPT2. This is perhaps an evidence of BERT model overfitting due to using a small classification head which is specifically designed for the downstream task (4 output nodes). 

% On the other hand, 
Since the language modeling loss benefits GPT2 model to exploit more supervision during training (predicting next token
% for words
% given the previous words
for all input tokens),
% in the input sequence), 
% \wl{the previous comment about drawing causal conclusion remain unaddressed}~\eh{please read my previous comment} \wl{how are we measuring the amount of supervision a model has? it doesn't have to be a quantitative measure, but it at least need to have a reasonable proxy} \wl{previous comment about measurement for supervison remains unaddressed} \eh{this is a known fact that language modeling loss exploits extra supervision than BERT, because of using all input positions to predict the next word, but BERT loss is only about predicting the class of CLS output} 
 perhaps this helps GPT2 to be less prune to overfitting, and outperforms BERT in few-shot setting.
% ~Figure~\ref{fig:ablation_convergence}(a) indicates that loss of label corresponding to the position of label prediction in GPT2 model, i.e. GPT2 (label position), converges faster than BERT model. This 
Additionally, reformulating the task as natural text might benefits GPT2 to infer the sentiment polarity easier than BERT. 
% the \textit{gpt2-generative (label position)} to converges faster is perhaps due to reformulating the task as natural text helps the GPT2 model to infer the sentiment polarity easier than BERT. 
Overall, GPT2 validation and test accuracy achieves higher performance. Analysis of training convergence on other tasks and domains are presented in Appendix~\ref{sec:appendix_ablation_training_convergence}, Figures~\ref{fig:appendix_ablation_convergence_semeval14} and~\ref{fig:appendix_ablation_convergence_semeval16}.

We also investigates model weights change during fine-tuning by measuring the average of the normalized weight update, Eq. (\ref{eq:mean_weight_update}),
% absolute value of weights changes (aggregate shift) 
for each layer (more details are presented in Appendix~\ref{sec:appendix_ablation_model_parameter} and Figure~\ref{fig:ablation_model_parameters}). It is shown that \textit{gpt2-generative} model has higher weight update in all layers at the end of training, and overall higher update in embedding layer (by one to two order of magnitude).
% the aggregate shift of GPT2 parameters are $\approx 1e-4$ during training, while BERT aggregate shift reduces to $\approx 1e-7$, with same pattern observed for self-attention layers too. 
% have similar pattern is observed. 
This observation perhaps indicates that standard language modeling loss provides more supervision to GPT2 model, when finetuned on few-shot data.

\vspace{-2mm}
\section{Conclusion}
\label{sec:conclusion}
\vspace{-2mm}
In this paper, we proposed 
% to use 
a generative language model for aspect based sentiment analysis (ABSA). By reformulating the task as language generation, the model learns to predict aspects and their 
% expressed opinion 
polarities
via language generation. Evaluation results on single-task polarity prediction on few and full shot setting indicate that the proposed approach outperforms prior arts, which are based on discriminative classification using BERT as encoder, with higher average performance and lower variance. On join-task 
% (aspect term or aspect category) 
and multi-task settings, 
% (aspect term and aspect category prediction),
the proposed model shows better performance on single-task polarity prediction metrics. Additionally, evaluation results on coarse-grained (SST2), fine-grained (SST5) sentiment analysis datasets, and OOS intent detection dataset indicate the better and more robust few-shot performance of generative language model. Furthermore,
qualitative analysis indicates that using 
% language generation in
multi-task setting improves model prediction via supervision across aspect term and category. 

\section{Broader Impact}
\label{sec:broader_impact}
This work may have implications for the simplification of sentiment analysis using neural text generation. 
In the narrow sense, this work addresses aspect-based sentiment analysis.
If so, the improvement of neural text generation systems and easier deployment would amplify both the positive and negative aspects of sentiment analysis.
On the positive side, neural text generation models might play a role in automating user opinion mining, and thereby increasing efficiency of currently modular systems.
On the negative side, it can dehumanize current systems, by automating systems towards multi-tasking, and reducing the level of human control on language generation. Moreover, this approach can introduce toxicity and biases into sentiment polarity predictions, such as gender, race, religious, and ethics~\cite{kiritchenko2018examining, park2018reducing}. This is due to biases which are learned during pretraining of neural text models on internet data~\cite{sheng2019woman, tan2019assessing}. 
%removing the need for human supervision.
These consequences are not specific to this work, but should be considered by the field of natural language processing more broadly.


\bibliography{sections/7-references}
\bibliographystyle{acl_natbib}

% \appendix

% \section{Example Appendix}
% \label{sec:appendix}

% This is an appendix.

\clearpage
\newpage
\newpage

\appendix

% \newpage
% \newpage

\section{Input Representation and Method Overview}
\label{appendix_input_rep}
As described in Section~\ref{subsec:trainig},
a single training sequence consists of the concatenation of review sentence $S^{k}$ with the corresponding aspect terms and their polarities $x^{k}=\left[S^{k}; T^{k}\right]$, or aspect categories and their polarities $x^{k}=\left[S^{k}; C^{k}\right]$. 

A schematic overview of each segment is shown in Table~\ref{tab:method} together with special tokens marking transition points. 
The generative language model is optimized by minimizing the negative likelihood over the joint sequence. 
% $x^t=[C_t; B_t; D_t; A_t; S_t]$. 
The output state associated with each input token is used to predict the next token.
During inference, for single task polarity prediction of each aspect term (sub-task SB1), the language model input comprises the review sentence concatenated by the corresponding aspect term. The the model generates a single token, which assumed as predicted polarity. Same method is used for sub-task SB4 for aspect category polarity prediction. For joint- and multi-task prediction, the input sequence contains only the review sentence. The language model then generates aspect terms and aspect categories along with their polarities in single toke-by-token generation, until the end-of-sentence special token is generated.

Examples of different input sequence formatting for different datasets evaluated in the paper are presented in~Table~\ref{tab:input_sequence_examples}. We are using identifiers to separate different segments of the input sequence. For example, to separate review sentence from aspect term, we introduced identifiers \textit{<|review|>} and \textit{<|term|>} to separate them. each segment also ends with an end-of-segment identifier, such as \textit{<|endofreview|>} and \textit{<|endofterm|>} identifiers. It is noteworthy that these identifiers are not special token, similar to BERT, which introduces new embeddings into vocabulary. We have noticed that defining identifiers as special token will decrease the performance of generative language model, perhaps due to introducing randomly-initialized embedding vectors into vocabulary, which requires more training data to finetune them. However, since GPT2 did not use special tokens during pretraining, using identifiers which are combination of pretrained vocabulary tokens and special characters, such as \{\textit{<, |, ,|, >}\}, helps GPT2 to understand different segments in the input sequence, to infer the sentiment polarity more accurately.

\begin{table*}[t!]
\small
\centering
\begin{tabular}{|m{10em}|m{30em}|}
\hline
Sentence $S^{k}$ & 
    \textcolor{blue}{[review]} 
    \textit{review sentence}
    \textcolor{blue}{[endofreview]}
    \\& \\
Aspect term $T^{k}$ &
 \textcolor{green}{[term]}
 $term_{1}$ $polarity_{1}$,  
 $term_{2}$ $polarity_{2}$,  
 $\ldots$
 $term_{I}$ $polarity_{I}$  
 \textcolor{green}{[endofterm]}
 \\
 & \\
Aspect category $C^{k}$ &
 \textcolor{purple}{[category]}
 $category_{1}$ $polarity_{1}$,  
 $category_{2}$ $polarity_{2}$,  
 $\ldots$
 $category_{J}$ $polarity_{J}$  
 \textcolor{purple}{[endofcategory]}
 \\
 & \\
Aspect term single and joint task training sequence ($LM_{term}$) & \textcolor{blue}{[review]} 
    \textit{review sentence}
    \textcolor{blue}{[endofreview]} \textcolor{green}{[term]}
 $term_{1}$ $polarity_{1}$,  
 $\ldots$
 \textcolor{green}{[endofterm]} \\
& \\
Aspect category single and joint task training sequence ($LM_{category}$) & \textcolor{blue}{[review]} 
    \textit{review sentence}
    \textcolor{blue}{[endofreview]} \textcolor{purple}{[category]}
 $category_{1}$ $polarity_{1}$,  
 $\ldots$
 \textcolor{purple}{[endofcategory]} \\
 & \\
Multi-task training sequence ($LM_{multi}$) & \textcolor{blue}{[review]} 
    \textit{review sentence}
    \textcolor{blue}{[endofreview]}
    \textcolor{green}{[term]}
 $term_{1}$ $polarity_{1}$,
 $\ldots$
 \textcolor{green}{[endofterm]}
 \textcolor{purple}{[category]}
 $category_{1}$ $polarity_{1}$,
 $\ldots$
 \textcolor{purple}{[endofcategory]} \\
 \\
\hline
\end{tabular}
\vspace{1mm}
\caption{A schematic representation of the different components of inputs/outputs in aspect-based sentiment analysis. When training generative language model, these are concatenated together into a single sequence, as shown in last three rows.}
\label{tab:method}
\end{table*}

\begin{table*}[htb!]
    \centering
    \small
    \begin{tabular}{ccp{5cm}p{5cm}}
    % \specialrule{.2em}{.2em}{.2em}
    \toprule[2pt]
     &  & \multicolumn{2}{c}{Aspect Category} \\
    \cline{3-4}
    Dataset & Domain & Entity & Attribute \\
    \midrule[1pt]
        \multirow{3}{*}{SemEval 14} &  \multirow{2}{*}{Restaurant} & ambience, anecdotes miscellaneous, food, price, service & N/A \\
        \cline{2-4}
         & Laptop & N/A & N/A \\
    \midrule[1pt]
    \multirow{8}{*}{SemEval 16} &  \multirow{2}{*}{Restaurant} & ambience, drinks, food, location, restaurant, service
    % ambience general, drinks prices, drinks quality, drinks style options, food prices, food quality, food style, location general, restaurant general, restaurant miscellaneous, restaurant prices, service general 
    & general, price, style, quality
    \\
        \cline{2-4}
         & \multirow{6}{*}{Laptop} & battery, company, cpu, display, fans cooling, graphics, hard disc, hardware, keyboard, laptop, memory, motherboard, mouse, multimedia devices, optical drives, os, ports, power supply, shipping, software, support, warranty
        & 
        miscellaneous, operation performance, quality, general, design features, usability, connectivity, portability, price 
        
        \\
        \specialrule{.2em}{.2em}{.2em}
    \end{tabular}
    \caption{Ascpet category definition for SemEval14 and SemEval16 datasets. In Semeval14, each unique aspect category is defined as entity. For SemEval16, aspect category is defined as combination of entity and attribute. Laptop domain does not have annotation in SemEval14 dataset.}
    \label{tab:appendix_aspect_category}
\end{table*}

\begin{table*}[h!]
    \centering
    \scriptsize
    \begin{tabular}{l|p{1.5cm}|p{1.5cm}|p{4cm}|p{4cm}}
    \specialrule{.2em}{.2em}{.2em}
         Dataset & Task & Type & \multicolumn{2}{c}{Input sequence} \\
         \cline{4-5}
         & & & train & inference \\
         \hline
        SemEval14 & Single task & aspect term polarity prediction
         & \textcolor{blue}{<|review|>} once we sailed, the top-notch food and live entertainment sold us on a unforgettable evening. \textcolor{blue}{<|endofreview|>} \textcolor{green}{<|term|>} food positive , live entertainment positive \textcolor{green}{<|endofterm|>} 
         & 
         \textcolor{blue}{<|review|>} once we sailed, the top-notch food and live entertainment sold us on a unforgettable evening. \textcolor{blue}{<|endofreview|>} \textcolor{green}{<|term|>} food \\
        % \hline
        \specialrule{.15em}{.2em}{.2em}
        SemEval14 & Joint task & aspect term 
         & \textcolor{blue}{<|review|>} once we sailed, the top-notch food and live entertainment sold us on a unforgettable evening. \textcolor{blue}{<|endofreview|>} \textcolor{green}{<|term|>} food positive , live entertainment positive \textcolor{green}{<|endofterm|>} 
         & 
         \textcolor{blue}{<|review|>} once we sailed, the top-notch food and live entertainment sold us on a unforgettable evening. \textcolor{blue}{<|endofreview|>}  \\
         \specialrule{.15em}{.2em}{.2em}
          SemEval14 & Multi-task & aspect term \& aspect category & \textcolor{blue}{<|review|>} the service was attentive without being overbearing and each dish we tried was wonderful from the spring rolls to the cod with pineapple tempura. \textcolor{blue}{<|endofreview|>} \textcolor{green}{<|term|>} service positive , dish positive , spring rolls positive , cod with pineapple tempura positive \textcolor{green}{<|endofterm|>} \textcolor{purple}{<|category|>} food positive , service positive \textcolor{purple}{<|endofcategory|>} 
          &
          \textcolor{blue}{<|review|>} the service was attentive without being overbearing and each dish we tried was wonderful from the spring rolls to the cod with pineapple tempura. \textcolor{blue}{<|endofreview|>}
          \\
        
        \specialrule{.15em}{.2em}{.2em}
         SST-2 & Single-task & polarity prediction & 
         \textcolor{blue}{<|review|>} does n't try to surprise us with plot twists , but rather seems to enjoy its own transparency \textcolor{blue}{<|endofreview|>} \textcolor{green}{<|sentiment|>} positive \textcolor{green}{<|endofsentiment|>} 
         &
         \textcolor{blue}{<|review|>} does n't try to surprise us with plot twists , but rather seems to enjoy its own transparency \textcolor{blue}{<|endofreview|>} \textcolor{green}{<|sentiment|>}
         \\
         
        \specialrule{.15em}{.2em}{.2em}
         SST-5 & Single-task & polarity prediction & \textcolor{blue}{<|review|>} it 's a lovely film with lovely performances by buy and accorsi . \textcolor{blue}{<|endofreview|>} \textcolor{green}{<|sentiment|>} somewhat positive \textcolor{green}{<|endofsentiment|>} 
         &
         \textcolor{blue}{<|review|>} it 's a lovely film with lovely performances by buy and accorsi . \textcolor{blue}{<|endofreview|>} \textcolor{green}{<|sentiment|>}
          \\
          
         \specialrule{.15em}{.2em}{.2em}
         OOS & Single-task & intent prediction & \textcolor{blue}{<|user|>} how would you say fly in italian \textcolor{blue}{<|endofuser|>} \textcolor{green}{<|intent|>} translate \textcolor{green}{<|endofintent|>} 
         &
         \textcolor{blue}{<|user|>} how would you say fly in italian \textcolor{blue}{<|endofuser|>} \textcolor{green}{<|intent|>}
         \\

        \specialrule{.2em}{.2em}{.2em}
        
    \end{tabular}
    \caption{Examples of input sequence during training and inference of generative language model for different datasets.
    }
    \label{tab:input_sequence_examples}
\end{table*}

\section{Multi-task prediction}
\label{sec:appendix_multitasking}

\begin{table*}[htb!]
    \centering
    \small
    \begin{tabular}{llrrrrr}
    \hline
    % \multirow{2}{*}{Tasks} &
    \multirow{2}{*}{Shot} & \multirow{2}{*}{Layers}& \multirow{2}{*}{Joint Accuracy} & \multicolumn{2}{c}{Term} & \multicolumn{2}{c}{Category} \\
    & & & SB1 (F1) & SB2 (Acc) & SB3 (F1) & SB4 (Acc)\\
    \hline
        %  \multirow{10}{*}{Multi} &
         \multirow{2}{*}{1\%} & 12 & 20.75 & 39.26 & 19.69 & 62.82 & 43.4 \\
          & 24 & 20.62 & 37.87 & 18.99 & 61.79 & 41.51 \\
        \cline{1-7}
         \multirow{2}{*}{5\%} & 12 & 31 & 44.35 & 32.38 & 74.46 & 56.51 \\
         & 24 & 34.87 & 60.4 & 35.18 & 75.39 & 59.06 \\
         \cline{1-7}
         \multirow{2}{*}{10\%} & 12 & 38.37 & 62.47 & 35.98 & 77.43 & 61.32 \\
         
         & 24 & 41.75 & 65.9 & 40.06 & 79.27 & 62.92 \\
         \cline{1-7}
         \multirow{2}{*}{20\%} & 12 & 42.88 & 66.82 & 39.91 & 79.39 & 62.36 \\
         
         & 24 & 45 & 72.73 & 45.31 & 80.79 & 65.28 \\
         \cline{1-7}
         \multirow{2}{*}{100\%} & 12 & 51.63 & 77.43 & 49.71 & 85.34 & 70.57 \\
         
         & 24 & 55.62 & 81.53 & 57.92 & 82.4 & 70.38 \\
         \cline{1-7}
         \hline
    \end{tabular}
    \caption{Multi-task evaluation on SemEval14 restaurant domain (SB1-4) on few-shot settings using generative language model (GPT2).}
    \vspace{-2mm}
    \label{tab:multi_task_semeval14_restaurant}
\end{table*}

\begin{table*}[h!]
    \centering
    \small
    \begin{tabular}{llrrrrr}
    \hline
    % \multirow{2}{*}{Tasks} &
    \multirow{2}{*}{Shot} & \multirow{2}{*}{Layers}& \multirow{2}{*}{Joint Accuracy} & \multicolumn{2}{c}{Term} & \multicolumn{2}{c}{Category} \\
    & & & SB1 (F1) & SB2 (Acc) & SB3 (F1) & SB4 (Acc)\\
    \hline
        %  \multirow{10}{*}{Multi} &
         \multirow{2}{*}{1\%} & 12 & 11.6 & 28.68 & 13.38 & 46.36 & 38.31 \\
          & 24 & 9.04 & 24.87 & 11.36 & 44.32 & 35.63 \\
        \cline{1-7}
         \multirow{2}{*}{5\%} & 12 & 18.43 & 33.81 & 16.74 & 56.85 & 50.06 \\
         & 24 & 20.48 & 34.99 & 18.88 & 61.09 & 54.66 \\
         \cline{1-7}
         \multirow{2}{*}{10\%} & 12 & 21.16 & 33.48 & 16.74 & 63.11 & 50.45 \\
         & 24 & 22.18 & 37.13 & 19.64 & 67.12 & 55.43 \\
         \cline{1-7}
         \multirow{2}{*}{20\%} & 12 & 25.77 & 37.74 & 20.63 & 69.39 & 62.07 \\
         & 24 & 26.96 & 40.6 & 22.15 & 72.9 & 65.39 \\
         \cline{1-7}
         \multirow{2}{*}{100\%} & 12 & 32.42 & 48.48 & 27.67 & 76.51 & 66.41 \\
         & 24 & 43 & 50.27 & 30.15 & 76.78 & 69.6 \\
         \cline{1-7}
         \hline
    \end{tabular}
    \caption{Multi-task evaluation on SemEval16 restaurant domain (SB1-4) on few-shot settings using generative language model (GPT2).}
    \label{tab:multi_task_semeval16_restaurant}
\end{table*}

In this section, evaluation results on SemEval 14 and SemEval16 restaurant domain are presented for multi-task learning using our proposed generative language model, based \textit{GPT2-base} model, in Tables ~\ref{tab:multi_task_semeval14_restaurant} and ~\ref{tab:multi_task_semeval16_restaurant}. For more details, please refer to section~\ref{exp:joint_task}.

\section{Ablation: Model input sequence formatting}
\label{sec:appendix_ablation_input_formatting_methods}
\begin{figure*}[htb!]
    \centering
    \subfigure[
    % semval14, restaurant term polarity
    ]{\includegraphics[width=0.3\linewidth]{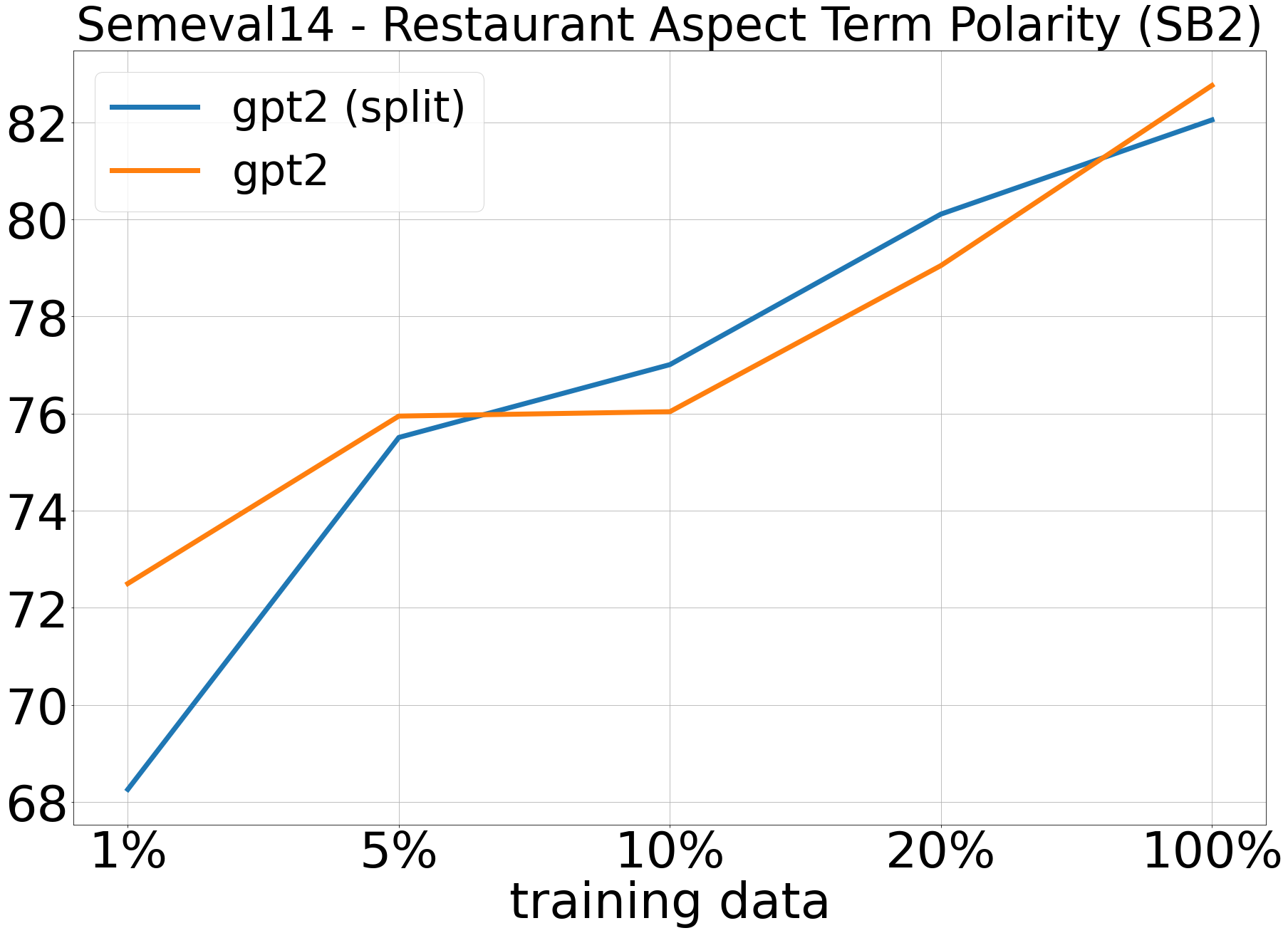}}
    \subfigure[
    % semval14, laptop term polarity
    ]{\includegraphics[width=0.3\linewidth]{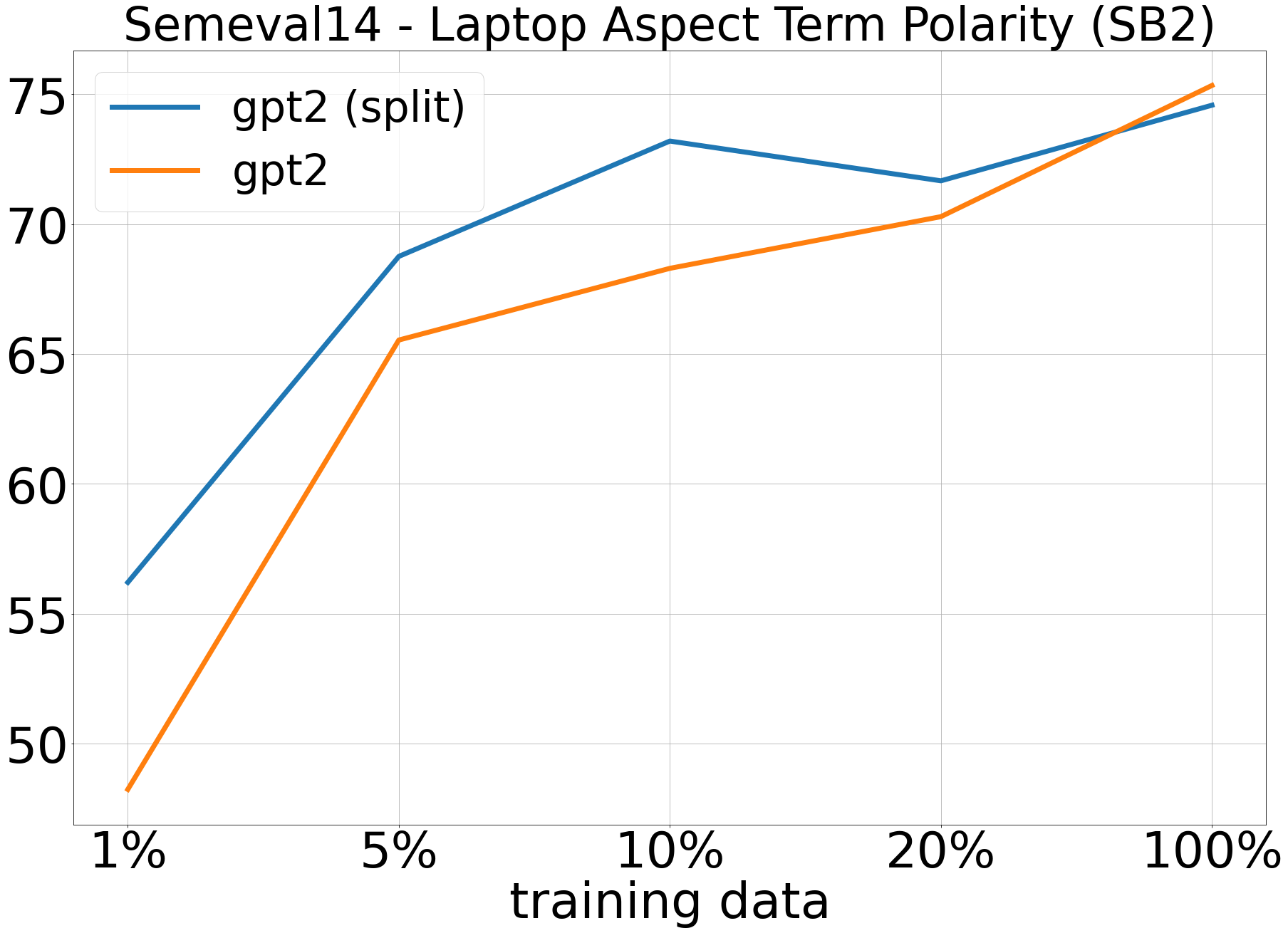}}
    \subfigure[
    % semval16, restaurant term polarity
    ]{\includegraphics[width=0.3\linewidth]{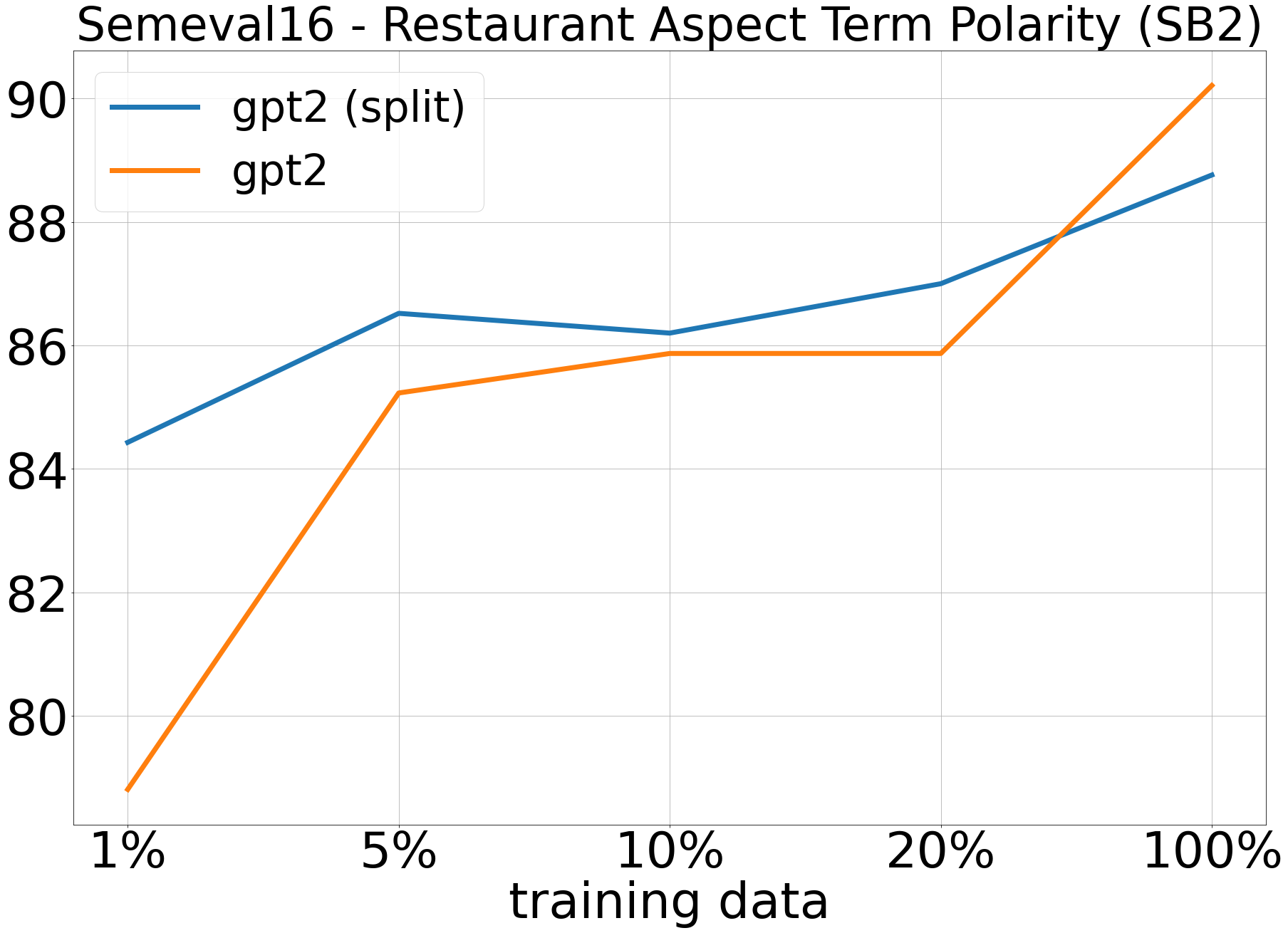}}
    % \vspace{-2mm}
    \caption{Ablation analysis on model input sequence formatting. \textit{GPT2 (split)} means review sentence is concatenated with each aspect terms separately. (best viewed in color)}
    \label{fig:appendix_ablation_input_format}
    % \vspace{-2mm}
\end{figure*}

% Increasing training data by splitting sequences improves few-shot performance:} 
For a single review sentence with multiple aspect terms or categories, there are two ways to create input sequence for language model training, as described in section~\ref{subsec:trainig}. First, the review sentence can be concatenated with each aspect terms separately (\textit{GPT2-Split}), which results in better performance for few-shot setting (Figure~\ref{fig:appendix_ablation_input_format}) 
% \wl{Figure 2 plot (a) seems to contradict this conclusion?} \eh{see highlighted text}. 
There are very few example in few-shot setting,
such as $20$ unique examples in $1\%$ setting, 
and using split method increases training data and perhaps mitigates model over-fitting. However, when the review sentence is concatenated with all pairs of aspect terms or categories in a single sequence, performance is better for full-shot setting. There are few exceptions in Figure~\ref{fig:appendix_ablation_input_format}(a) for 1\% and 5\% shot settings. We observe that 1\% few-shot contains 20, 14, 12 input sequences in Figure~\ref{fig:appendix_ablation_input_format}(a), (b), and (c), respectively, for the regular method. However, the split method increases input training sequences to 36, 23, 17. It means that when the number of training sequences are high enough, increasing number of training examples using split methods might deteriorates the few-shot performance, as shown in Figure~\ref{fig:appendix_ablation_input_format}(a). 
We guess that the better few-shot performance of the \textit{GPT2-Split} method possibly depends on the number of unique training sequences when comparing to the regular method. In other words, the \textit{GPT2-Split} methods might outperforms the regular method when the number of training sequences is very low.

\section{Ablation: Generative vs. Discriminative language model}
\label{sec:appendix_ablation_generative}

In this section, ablation analysis on using generative language model as a classifier are presented in
Figures~\ref{fig:appendix_ablation_convergence_semeval14} and~\ref{fig:appendix_ablation_convergence_semeval16}.
% Figure~\ref{fig:ablation_generative}. 
It is shown that when fine-tuning \textit{GPT2} model as a classifier on the downstream task using an classification layer, it under-performs BERT model on few and full-shot settings. For more details, please refer to section~\ref{exp:ablation}.

\section{Ablation: Training convergence}
\label{sec:appendix_ablation_training_convergence}
\begin{figure*}[htb!]
    \centering

    % \subfigure[SemEval14 Restaurant Aspect Term Polarity (SB2)
    % ]{\includegraphics[width=0.7\linewidth]{figures/ablation/training_convergence/semeval14_restaurant_term_train_val.png}\label{subfig2}}
    
    \subfigure[SemEval14 Laptop Aspect Term Polarity (SB2)
    ]{\includegraphics[width=0.9\linewidth]{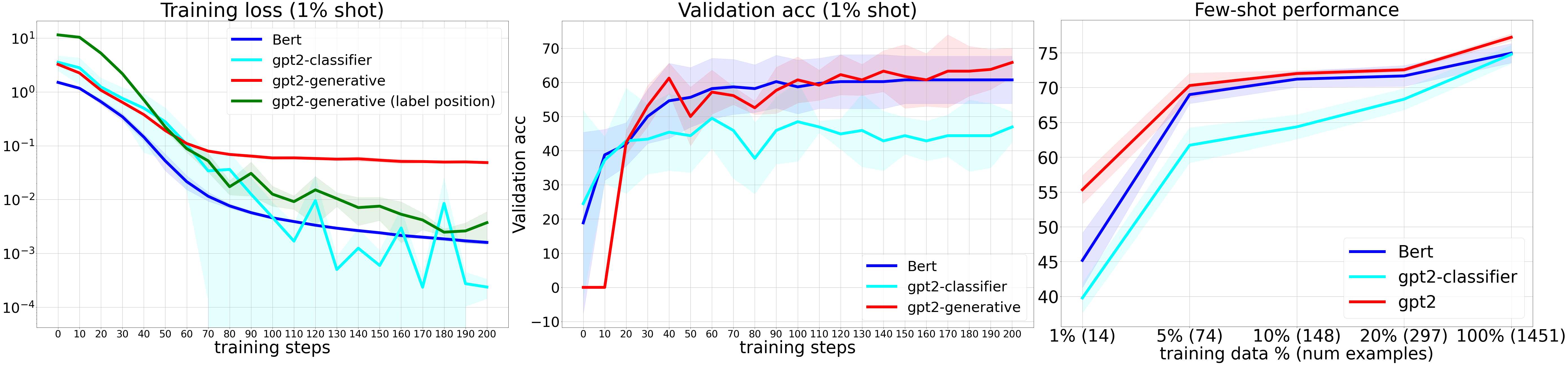}\label{subfig2}}
    
    \subfigure[SemEval14 Restaurant Aspect Category Polarity (SB4)
    ]{\includegraphics[width=0.9\linewidth]{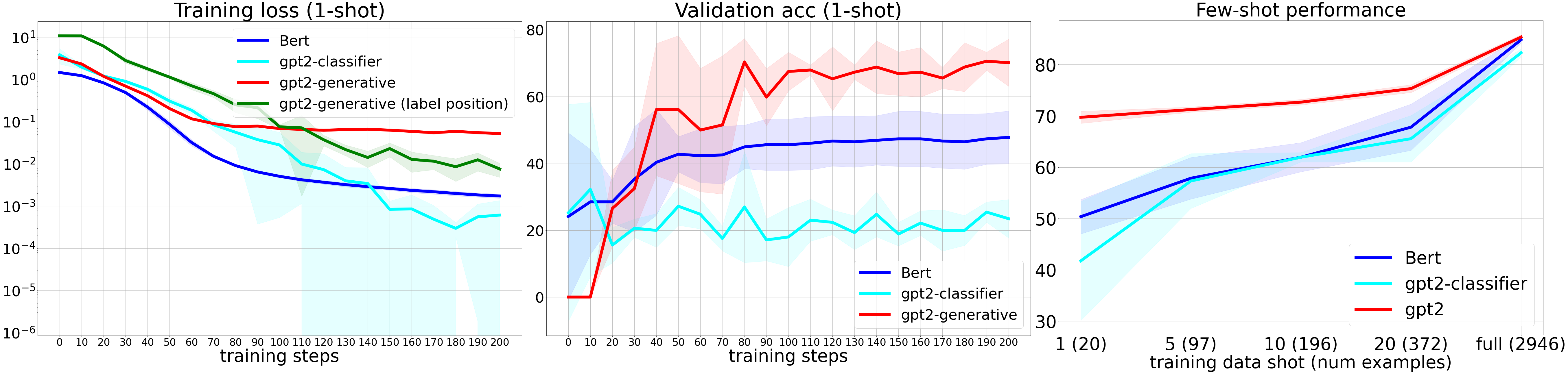}}

    \caption{Analysis of few-shot training convergence, evaluated on SemEval14 for $1\%$ and $1$-shot training data, and few-shot performance on all settings (right). GPT2-classifier model uses a classification layer on the output of last input token without using language modeling loss for training.
    (best viewed in color)}
    \label{fig:appendix_ablation_convergence_semeval14}
    % \vspace{-2mm}
\end{figure*}

\begin{figure*}[htb!]
    \centering

    \subfigure[SemEval16 Restaurant Aspect Term Polarity (SB2)
    ]{\includegraphics[width=0.9\linewidth]{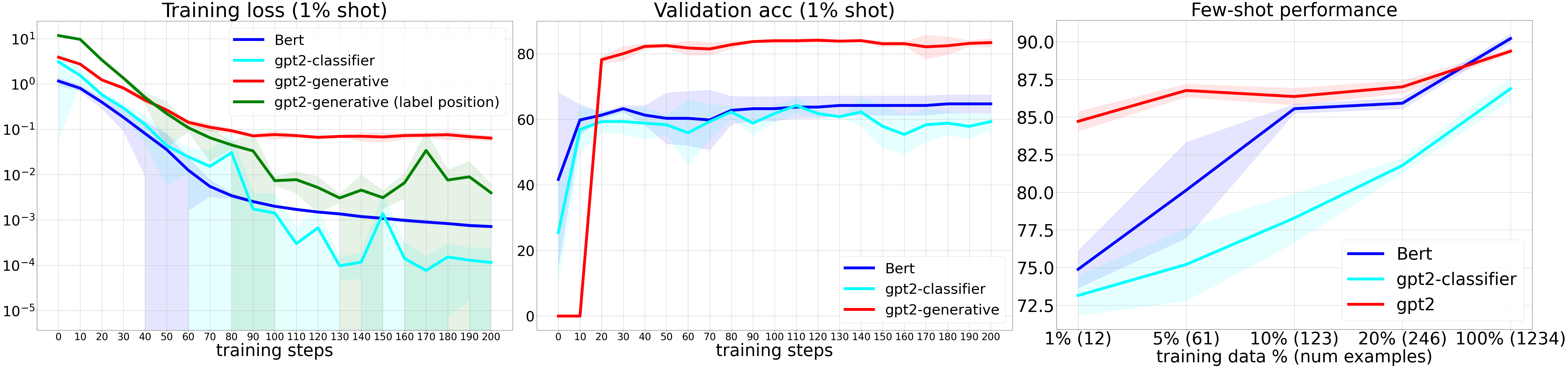}\label{subfig2}}
    
    \subfigure[SemEval16 Restaurant Aspect Category Polarity (SB4)
    ]{\includegraphics[width=0.9\linewidth]{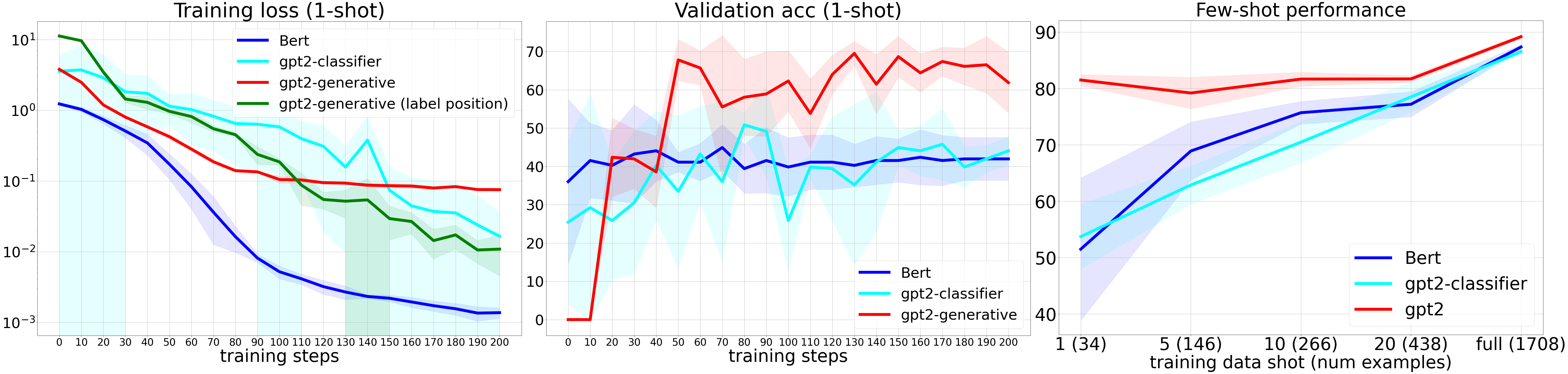}\label{subfig2}}
    
    \subfigure[SemEval16 Laptop Aspect Category Polarity (SB4)
    ]{\includegraphics[width=0.9\linewidth]{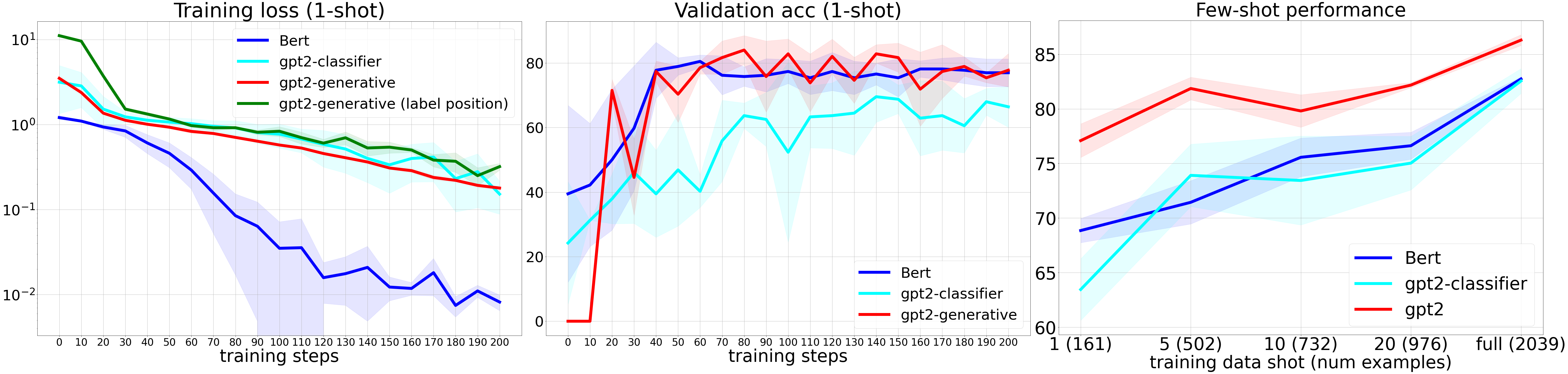}\label{subfig2}}
    
    \caption{Analysis of few-shot training convergence, evaluated on SemEval16 for $1\%$ and $1$-shot training data, and few-shot performance on all settings (right). GPT2-classifier model uses a classification layer on the output of last input token without using language modeling loss for training.
    (best viewed in color)}
    \label{fig:appendix_ablation_convergence_semeval16}
    % \vspace{-2mm}
\end{figure*}

In this section, training convergence of GPT2 model is compared with BERT and GPT2-classifier model in varios tasks of aspect-based sentiment analysis. As shown in Figures~\ref{fig:appendix_ablation_convergence_semeval14}
 and~\ref{fig:appendix_ablation_convergence_semeval16}, GPT2 achieves higher validation accuracy, when its training losses, standard language modeling and loss corresponding to label position, have higher value than BERT and GPT2-classifier. This indicates that perhaps BERT and GPT2-classifier overfitted to the few-shot training data. On the other hand, GPT2 language model achieves more supervision via standard language modeling loss, which results in higher training loss, but better validation performance.
 
\section{Ablation: Model weights update during training}
\label{sec:appendix_ablation_model_parameter}
\begin{figure*}[htb!]
    \centering
    \subfigure[
    ]{\includegraphics[width=0.45\linewidth]{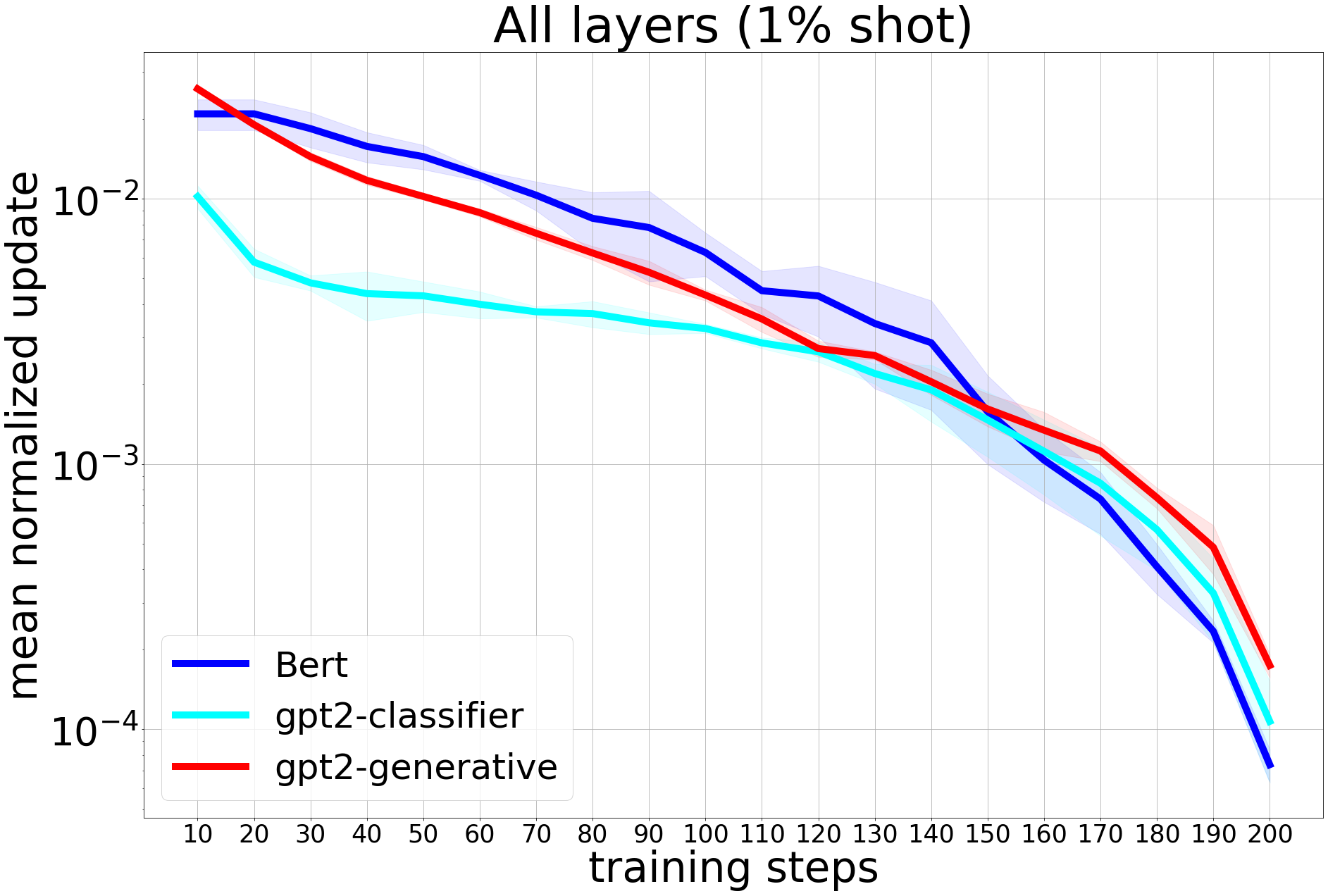}}
    \subfigure[
    ]{\includegraphics[width=0.45\linewidth]{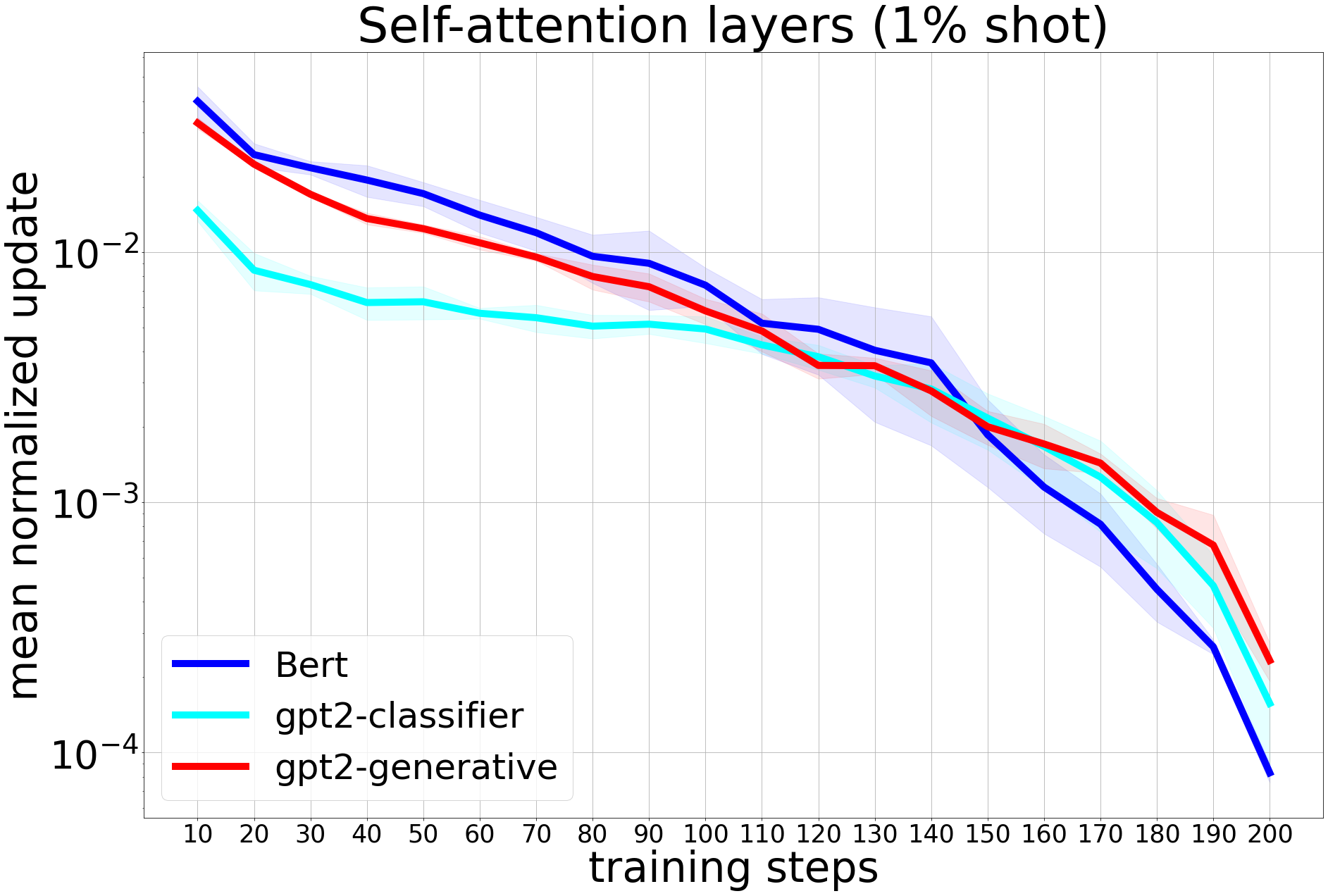}}
    
    \subfigure[
    ]{\includegraphics[width=0.45\linewidth]{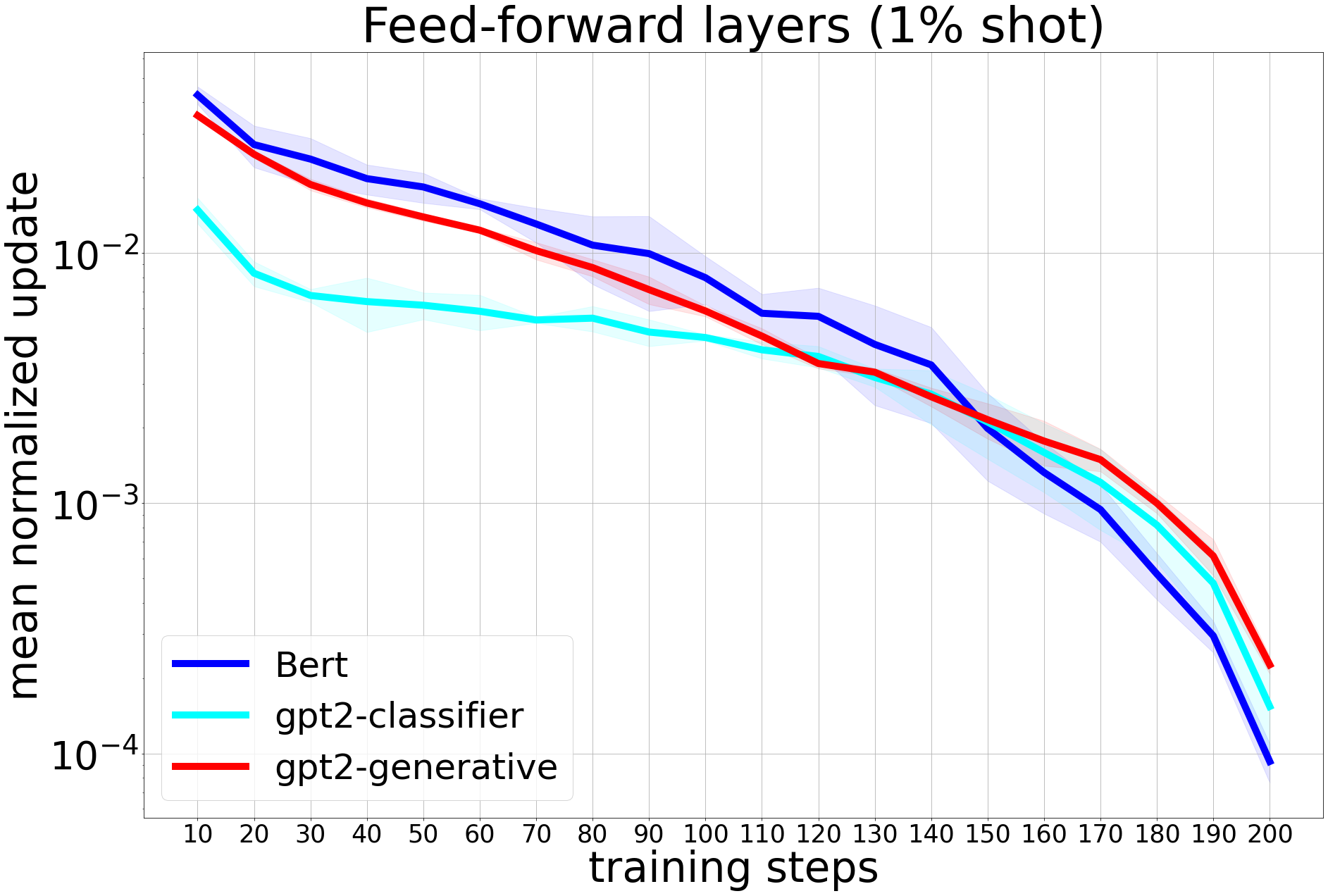}}
    \subfigure[
    ]{\includegraphics[width=0.45\linewidth]{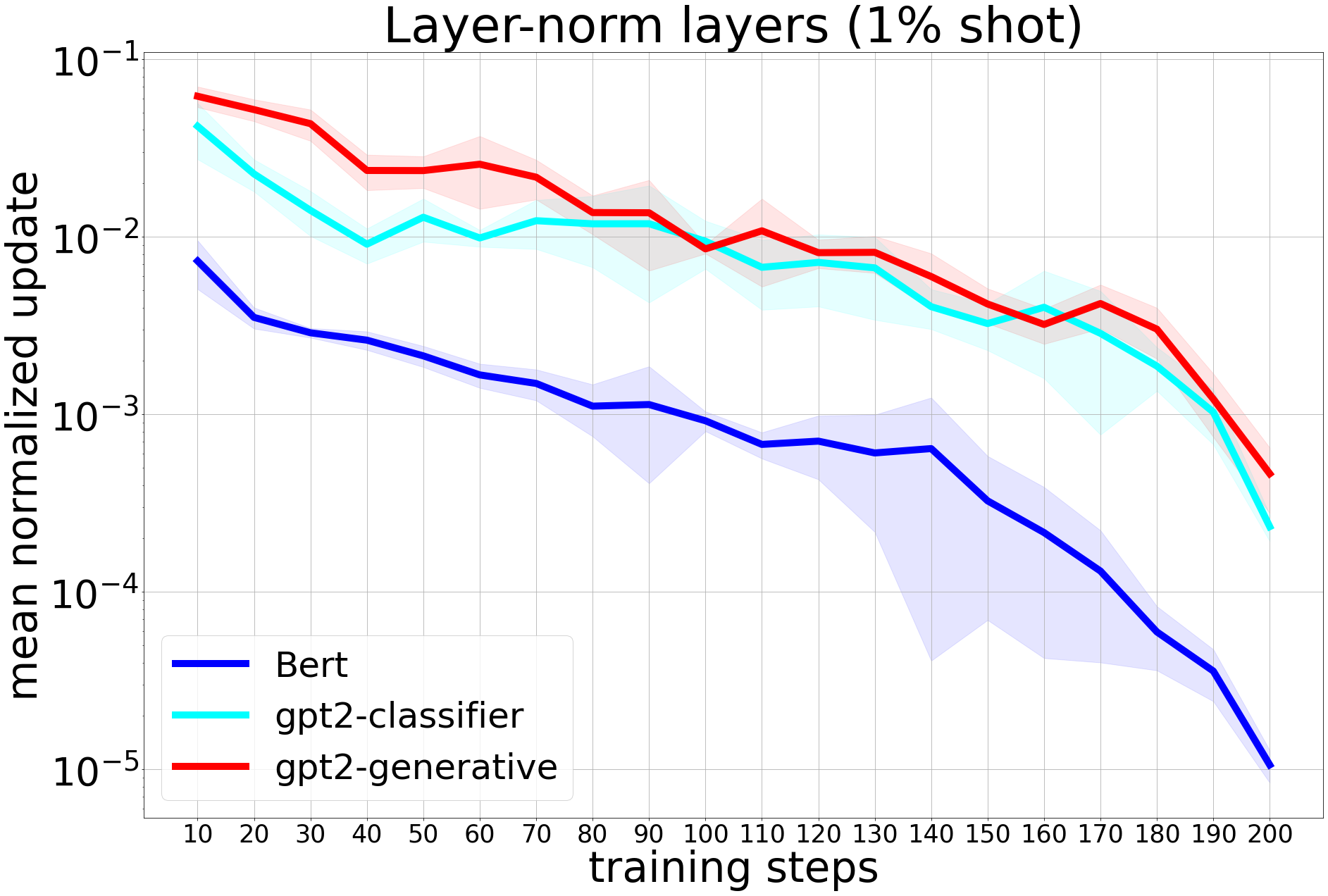}}
    
    \subfigure[
    ]{\includegraphics[width=0.45\linewidth]{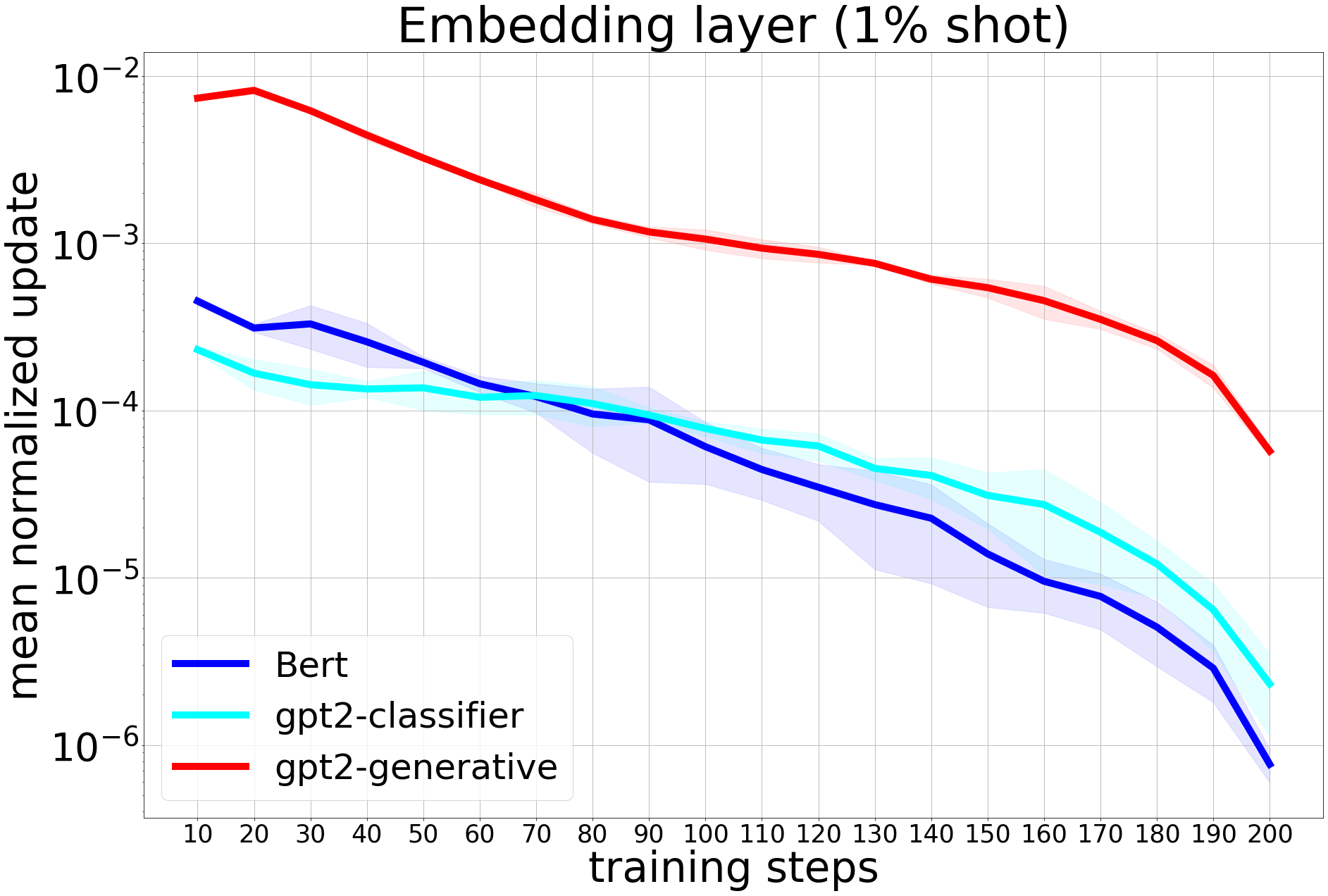}}
    \subfigure[
    ]{\includegraphics[width=0.45\linewidth]{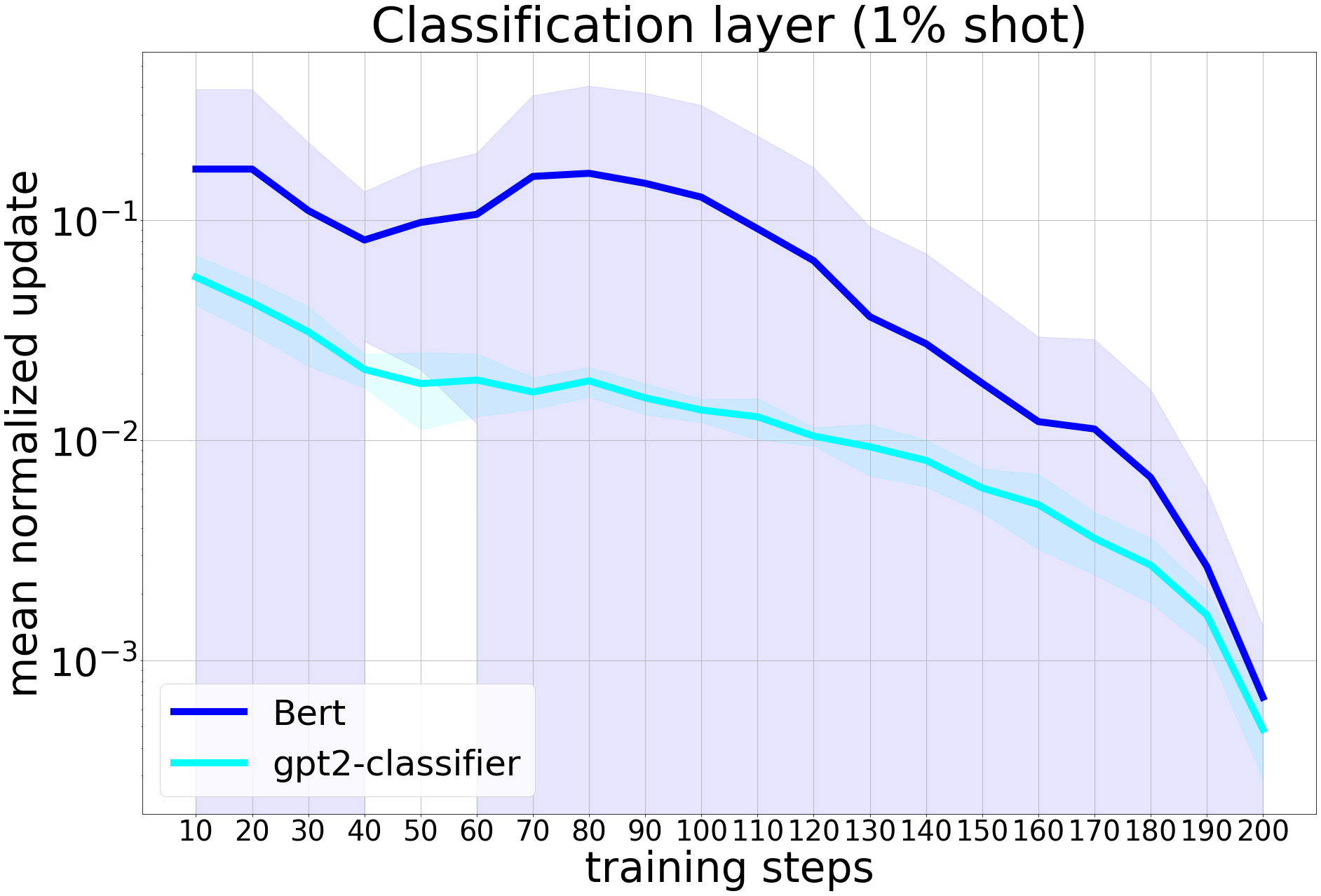}}
    
    % \subfigure[
    % ]{\includegraphics[width=0.45\linewidth]{figures/gpt2_all_weights_sum.png}}
    % \subfigure[
    % ]{\includegraphics[width=0.44\linewidth]{figures/gpt2_attention_sum.png}}
    % \vspace{-2mm}
    \caption{Model Layers mean normalized update, Eq. (~\ref{eq:mean_weight_update}) , during training. Normalized update of the weight $w$ at training step $i$ is defined as $(w_{i} - w_{i-1})/w_{0}$.
    % for all weights in that layer.
    Results are for training on 1\% few-shot data on SemEval14 restaurant aspect term polarity (SB2) prediction task for4 random seed. Shaded area indicates standard deviation.}
    \label{fig:ablation_model_parameters}
    % \vspace{-2mm}
\end{figure*}

In order to understand models behavior during training on few-shot data, we study the weight update at each layer of GPT2 and BERT models, during training on 1\% few-shot data. For each layer, the mean normalized weight update is defined as, 
\begin{equation}
\sum_{i=0}^{k} \frac{(w^{l}_{i} - w^{l}_{i-1})}{w^{l}_{0}}
\label{eq:mean_weight_update}
\end{equation}
where $l$ indicate the layer index, $i$ indicates training step, and $w^{l}_{0}$ refers to initial weight value before training.
% The aggregate shift is computed by the mean value of the absolute value of weight change for each parameter $|w_{i+1} - w_{i}|$.  
The comparison between GPT2 as generative \textit{gpt2-generative}, GPT2 as an ecoder for classification \textit{gpt2-classifie} and BERT model when trained on $1\%$ few shot data of SemEval14 restaurant domain are shown in Figure~\ref{fig:ablation_model_parameters}. The results indicate that Bert model has higher variance for all layers, especially for the randomly-initialized classification layer. Moreover, the mean normalized update of BERT model is larger that \textit{gpt2-generative} early during training, but is smaller at the end of training, where \textit{gpt2-generative} achieves higher validation performance, as shown in Figure~\ref{fig:ablation_convergence}. Furthermore, the mean normalized update in embedding layer of \textit{gpt2-generative} is significantly larger than \textit{BERT} and \textit{gpt2-classifier} by one order of magnitude early at training, which increased to two order of magnitude at the end. We conjecture that higher value in layer weights update at embedding layers, and at the end of training for other layers is perhaps due to using standard language modeling loss, which may provide more supervision signal for GPT2, compared to cross-entropy loss in BERT and \textit{gpt2-classifier} models.

\section{Ablation: Other Sentiment Analysis Tasks}
\label{sec:appendix_other_sentiment}
\begin{figure*}[htb!]
    \centering
    % \subfigure[
    % ]{\includegraphics[width=0.6\linewidth]{figures/sst_intent/sst2.png}}
    
    \subfigure[
    ]{\includegraphics[width=0.6\linewidth]{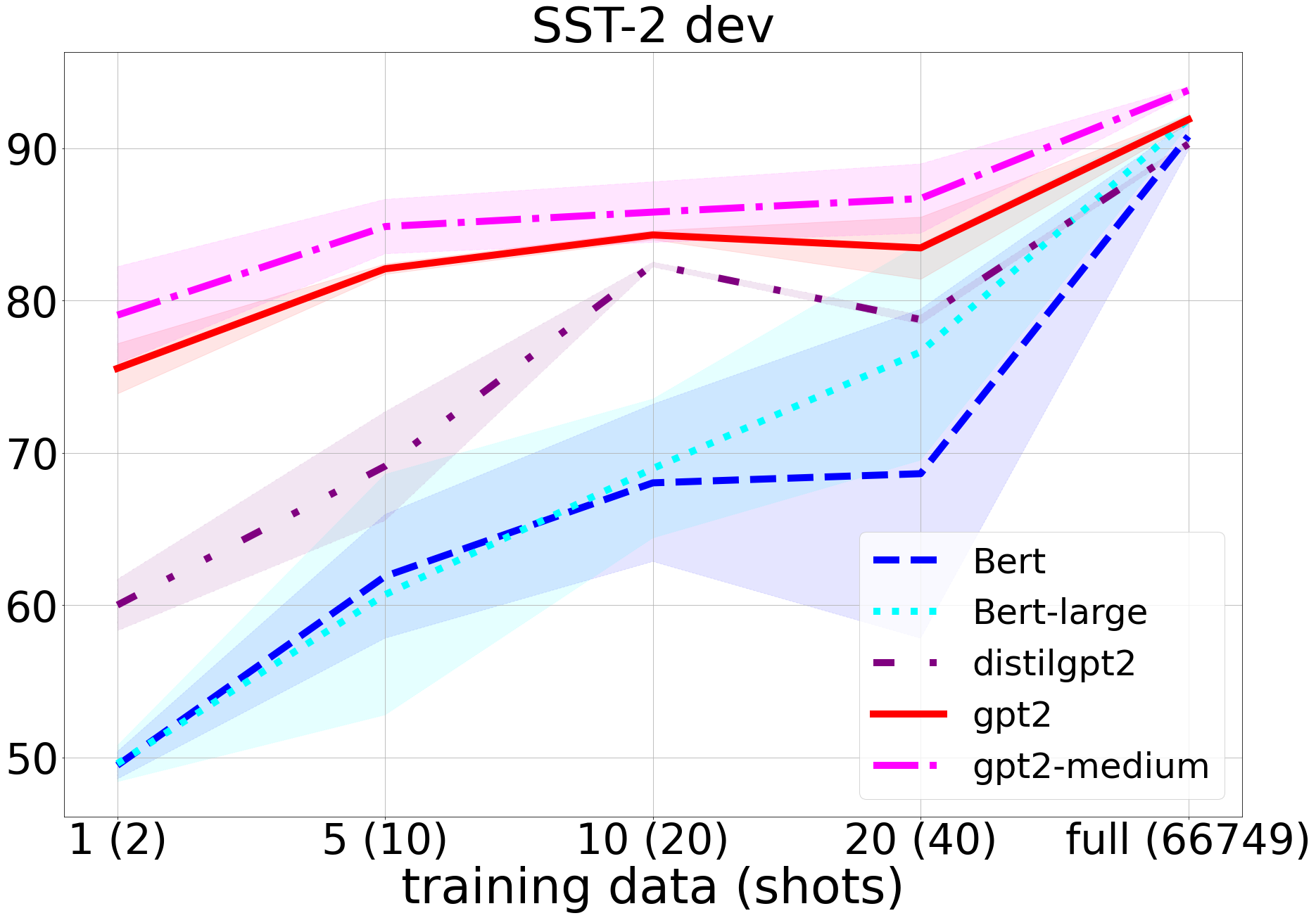}}
    
    \subfigure[
    ]{\includegraphics[width=0.6\linewidth]{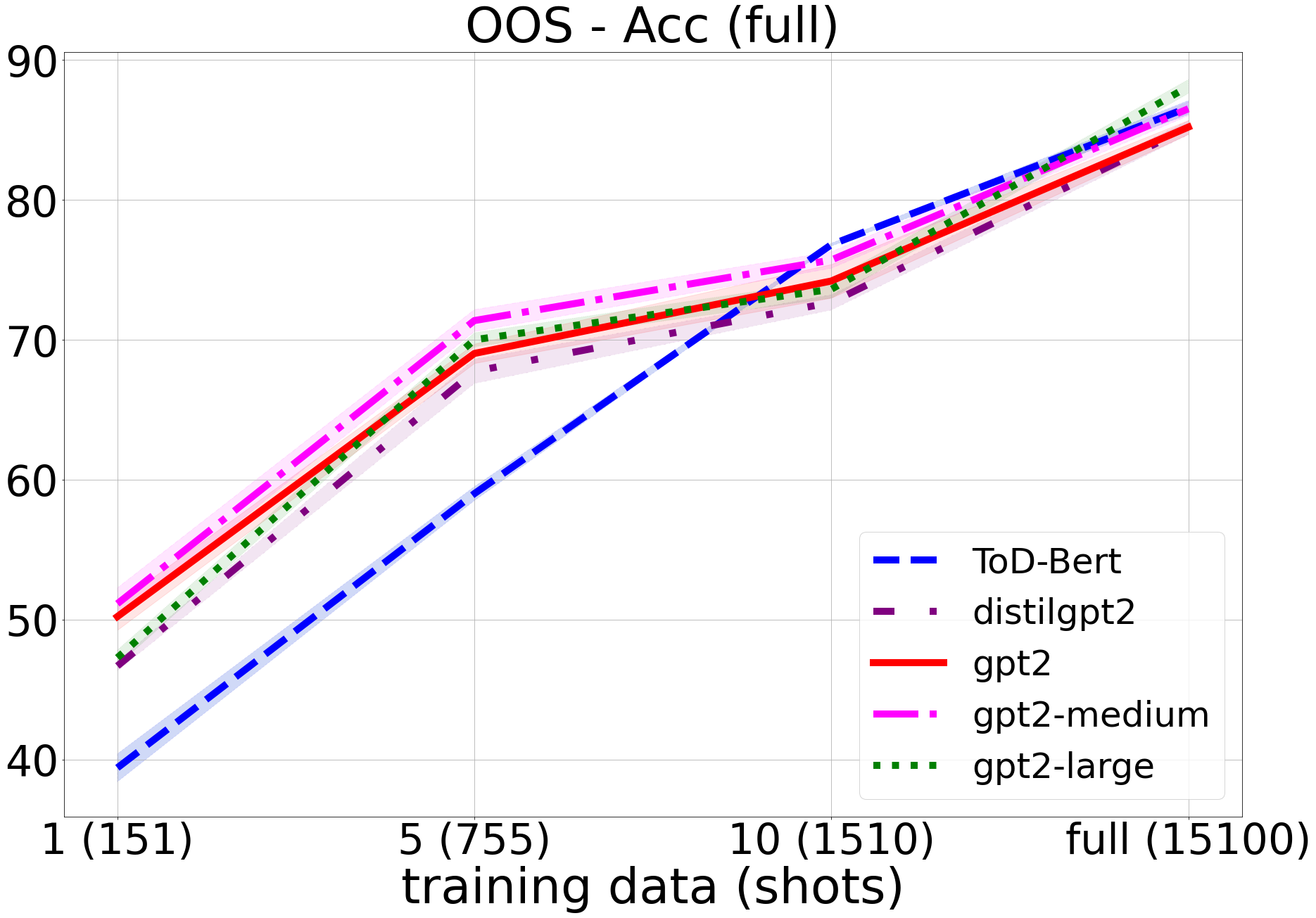}}
    % \vspace{-2mm}
    \caption{Few-shot evaluation of GPT2 and BERT models on SST2 dev set and OOS intent detection datasets. Note: 1-shot refers to one example per class. 
    % For aspect term, the percentage of training data are used for few shot settings. 
    (best viewed in color) 
    }
    \label{fig:ablation_other_sentiment_appendix}
    % \vspace{-2mm}
\end{figure*}

% \paragraph{Generative language model (GPT2) outperforms BERT on sentiment analysis without extra pretraining:} 
In order to extend the investigate the performance of our proposed generative language model to other sentiment analysis tasks, we also evaluate few-shot performance on SST-5  sentiment analysis dataset~\cite{socher-etal-2013-recursive} (binary and fine-grained sentiment classification), and OOS~\cite{larson-etal-2019-evaluation} intent detection dataset. The results are shown in Figure~\ref{fig:ablation_other_sentiment_appendix}, which indicate the superiority of generative model (GPT2) over discriminative BERT. On intent detection,~Figure~\ref{fig:ablation_other_sentiment_appendix}(c), GPT2 also outperforms TOD-BERT~\cite{wu-etal-2020-tod} which exploits extra pretraining on dialogue datasets to increase its few-shot performance.

\section{Qualitative Analysis}
\label{exp:analysis}

As described in section~\ref{exp:joint_task} and Table~\ref{tab:joint_task}, aspect term extraction on restaurant domain (SemEval14) is improved in multi-task learning. To better understand model behavior, some examples are shown in Table~\ref{tab:qualitative_positive}. 
Using aspect category as supervision in multi-task learning helps the model to more accurately generates the aspect terms, reduces false positive aspect terms and wrong polarity predictions. Moreover, multi-tasking helps to better predict category polarity, using supervision from aspect term during training. Some examples of wrong prediction are shown in Table~\ref{tab:qualitative_worng_pred}. It indicates that when there are negative or conflict polarity, the model struggles to correctly predict everything correctly. 
This often happens when there are opposite opinions for different aspect terms or categories.

% \section{Qualitative Analysis}
% \label{sec:appendix_qualitative}
% In this section, qualitative analysis of our proposed generative language model predictions are presented. As shown in Table~\ref{tab:qualitative_worng_pred}, examples of wrong predictions are presented for joint and multi-task learning. It indicates that when there are contradicting opinions on multiple aspect terms and categories, especially in case of \verb|neutral| and \verb|conflict|, the model has difficulty in correctly predicting all polarities. For more discussion, please refer to section~\ref{exp:analysis}.

\begin{table*}[htb!]
    \centering
    \scriptsize
    \begin{tabular}{p{6cm}|l|p{5cm}}
    \specialrule{.2em}{.2em}{.2em}
         Sentence & Task & Model Output \\
        %  \hline
         \specialrule{.15em}{.2em}{.2em}
        the sangria's - watered down. & 
        % term polarity & sangria negative \\
        % \cline{2-3}
        aspect term & <|term|> sangria negative \\
        \cline{2-3}
        % & category polarity & food negative\\
        % \cline{2-3}
        & aspect category & <|category|> food \color{red}{neutral} \\
        \cline{2-3}
        & aspect term \& category & <|term|> sangria negative <|category|> food \color{green}{negative}\\
        \cline{2-3}
        & groundtruth & <|term|> sangria negative <|category|> food negative \\
        % \hline
        \specialrule{.15em}{.2em}{.2em}
        everyone who works there (the host, the bartender, the servers) is so helpful. & aspect term & host positive, bartender \textcolor{red}{neutral}, servers positive \\
        \cline{2-3}
        & aspect category & <|category|> service positive \\
        \cline{2-3}
        & aspect term \& category & <|term|> host positive, bartender \textcolor{green}{positive}, servers positive <|category|> service positive \\
        \cline{2-3}
        & groundtruth & <|term|> bartender positive, host positive, servers positive <|category|> service positive \\
        % \hline
        \specialrule{.15em}{.2em}{.2em}
        in mi burrito, here was nothing but dark chicken that had that cooked last week and just warmed up in a microwave taste. & aspect term & <|term|> \textcolor{red}{dark chicken} negative, \textcolor{red}{microwave taste neutral} \\
        \cline{2-3}
        & aspect category & <|category|> food negative \\
        \cline{2-3}
        & aspect term \& category & <|term|> \textcolor{green}{chicken} negative, \textcolor{green}{taste} negative <|category|> food negative\\
        % \hline
        \specialrule{.15em}{.2em}{.2em}
        if you like seafood and or greek food you will love this place though it is not limited to just these things. & aspect term & <|term|> seafood positive, greek food positive, \textcolor{red}{place negative} \\
        \cline{2-3}
        & aspect category & <|category|> food positive \\
        \cline{2-3}
        & aspect term \& category & <|term|> seafood positive, greek food positive <|category|> food positive\\
        \cline{2-3}
        & groundtruth & <|term|> greek food positive, seafood positive <|category|> food positive \\
        % \hline
        \specialrule{.2em}{.2em}{.2em}
        
    \end{tabular}
    \caption{Examples of correct predictions in multi-task learning.
    }
    \label{tab:qualitative_positive}
    \vspace{-2mm}
\end{table*}

\begin{table*}[h!]
    \centering
    \scriptsize
    \begin{tabular}{p{6cm}|l|p{5cm}}
    \specialrule{.2em}{.2em}{.2em}
         Sentence & Task & Model Output \\
         \hline
        certainly not the best sushi in new york, however, it is always fresh, and the place is very clean, sterile.
         & 
        aspect term & <|term|> sushi \textcolor{red}{negative}, place positive \\
        \cline{2-3}
        & aspect category & <|category|> ambience positive, food \textcolor{red}{positive} \\
        \cline{2-3}
        & aspect term \& category & <|term|> sushi \textcolor{red}{positive}, place positive <|category|> food \textcolor{red}{positive}, ambience positive  \\
        \cline{2-3}
        & groundtruth & <|term|> place positive, sushi conflict <|category|> ambience positive, food conflict \\
        % \hline
        \specialrule{.15em}{.2em}{.2em}
        while there's a decent menu, it shouldn't take ten minutes to get your drinks and 45 for a dessert pizza. & aspect term & menu positive, drinks \textcolor{red}{positive}, dessert pizza \textcolor{red}{positive} \\
        \cline{2-3}
        & aspect category & food \textcolor{red}{conflict} \\
        \cline{2-3}
        & aspect term \& category & <|term|> menu positive, drinks \textcolor{red}{positive}, dessert pizza \textcolor{red}{positive} <|category|> food \textcolor{green}{positive} \\
        \cline{2-3}
        & groundtruth & <|term|> dessert pizza neutral, drinks neutral, menu positive <|category|> food positive, service negative \\
        % \hline
        \specialrule{.15em}{.2em}{.2em}
        the portions of the food that came out were mediocre. & aspect term & \textcolor{red}{portions negative, food} neutral \\
        \cline{2-3}
        & aspect category & food \textcolor{red}{negative} \\
        \cline{2-3}
        & aspect term \& category & <|term|> \textcolor{red}{portions negative, food negative} <|category|> food \textcolor{red}{negative} \\
        \cline{2-3}
        & groundtruth & <|term|> portions of the food neutral <|category|> food neutral \\
        % \hline
        \specialrule{.2em}{.2em}{.2em}
        
    \end{tabular}
    \caption{Examples of wrong prediction for joint and multi-task generative language model.
    }
    \label{tab:qualitative_worng_pred}
\end{table*}

\end{document}